\documentclass{article}
\usepackage[preprint]{arxiv}
\usepackage[hidelinks]{hyperref}        
\usepackage{booktabs}                   
\usepackage{microtype}                  
\usepackage{xcolor}                     
\usepackage{doi}
\usepackage{authblk}
\usepackage[english]{babel}
\usepackage{amsfonts}       
\usepackage{amssymb}        
\usepackage{mathtools}      
\usepackage{nicefrac}       
\usepackage{cleveref}
\usepackage{glossaries-extra}
\setabbreviationstyle[acronym]{long-short}
\usepackage{graphicx}
\usepackage[font=small]{caption}
\usepackage{subcaption}
\usepackage[section]{placeins}
\usepackage{floatrow}
\usepackage{algorithm}
\usepackage{algpseudocodex}
\usepackage{siunitx}

\newacronym{dl}{DL}{deep learning}
\newacronym{psnr}{PSNR}{peak signal-to-noise ratio}
\newacronym{ssim}{SSIM}{structural similarity index measure}
\newacronym{mse}{MSE}{mean squared error}
\newacronym{cnn}{CNN}{convolutional neural network}
\newacronym{nn}{NN}{neural network}
\newacronym{cv}{CV}{computer vision}
\newacronym{icl}{ICL}{in-context learning}
\newacronym{llm}{LLM}{large language model}
\newacronym{cdp}{CDP}{common depth point}
\newacronym{nmo}{NMO}{normal moveout}
\newacronym{leakyrelu}{LeakyReLU}{leaky rectified linear function}
\newacronym{srme}{SRME}{Surface-Related Multiple Elimination}
\newacronym{rt}{RT}{Radon transform}

\newcommand{\unet}{U-Net}
\newcommand{\universeg}{UniverSeg}
\newcommand{\model}{ContextSeisNet}

\newcommand{\notationXm}{\mathbf{X}[m]}
\newcommand{\notationVm}{\mathcal{V}_m}
\newcommand{\notationYm}{\mathbf{Y}^*_m}

\title{In-Context Learning for Seismic Data Processing}

\newbox{\orcid}\sbox{\orcid}{\includegraphics[scale=0.06]{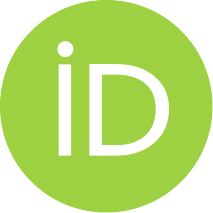}}
\author[1,2]{%
  \href{https://orcid.org/0009-0009-1060-6093}{\usebox{\orcid}\hspace{1mm}Fabian Fuchs\thanks{\texttt{fabian.fuchs@itwm.fraunhofer.de}}}
}
\author[1]{%
  \href{https://orcid.org/0009-0006-3945-1863}{\usebox{\orcid}\hspace{1mm}Mario Ruben Fernandez}%
}
\author[1]{%
  Norman Ettrich
}
\author[2,3]{%
  \href{https://orcid.org/0000-0002-1327-1243}{\usebox{\orcid}\hspace{1mm}Janis Keuper}%
}
\affil[1]{Fraunhofer-Institut für Techno- und Wirtschaftsmathematik}
\affil[2]{DWS, University of Mannheim}
\affil[3]{IMLA, Offenburg University}

\begin{document}

\maketitle

\begin{abstract}

    Seismic processing transforms raw data into subsurface images essential for geophysical applications. 
    Traditional methods face challenges, such as noisy data, and manual parameter tuning, among others.
    Recently deep learning approaches have proposed alternative solutions to some of these problems.
    However, important challenges of existing deep learning approaches are spatially inconsistent results across neighboring seismic gathers and lack of user-control. 

    We address these limitations by introducing \model{}, an \glsxtrlong{icl} model, to seismic demultiple processing. 
    Our approach conditions predictions on a support set of spatially related example pairs: neighboring common-depth point gathers from the same seismic line and their corresponding labels. 
    This allows the model to learn task-specific processing behavior at inference time by observing how similar gathers should be processed, without any retraining. 
    This method provides both flexibility through user-defined examples and improved lateral consistency across seismic lines.
    
    On synthetic data, \model{} outperforms a U-Net baseline quantitatively and demonstrates enhanced spatial coherence between neighboring gathers. 
    On field data, our model achieves superior lateral consistency compared to both traditional Radon demultiple and the U-Net baseline. 
    Relative to the U-Net, \model{} also delivers improved near-offset performance and more complete multiple removal. 
    Notably, \model{} achieves comparable field data performance despite being trained on 90\% less data, demonstrating substantial data efficiency.
    
    These results establish  \model{} as a practical approach for spatially consistent seismic demultiple with potential applicability to other seismic processing tasks.
\end{abstract}

\section{Introduction}
\label{sec:introduction}
Converting raw seismic data into interpretable subsurface images is critical for geophysical analysis.
Accurate subsurface imaging underpins structural interpretation, petroleum exploration, reservoir characterization, and geothermal studies \cite[]{yilmaz_seismic_2001-1}.
However, this conversion faces several challenges, such as ambient noise interference, sensor failures, and diminished low-frequency content, all of which degrade data quality.
To address these issues, traditional seismic processing relies on different sets of algorithms, tailored to each step in the workflow.
These conventional methods, however, require manual, iterative parameter selection (e.g., velocity picking, mute function) tailored to each dataset, and often rely on approximations of wave propagation.
Consequently, effective seismic processing demands substantial specialist expertise to recover subsurface information while minimizing acquisition and processing artifacts.

Within this broader processing workflow, multiple attenuation plays a pivotal role by enhancing migration results and enabling clearer geological interpretation.
Demultiple procedures typically precede velocity analysis and exploit the periodicity and predictability of multiples without requiring a velocity model.
A common approach is \gls{srme} \cite[]{verschuur_adaptive_1992}), which leverages the fact that surface multiples can be represented as combinations of primary raypaths to construct a multiple model \cite[]{verschuur_seismic_2013}.
After predicting the multiples, adaptive subtraction is applied to remove them from the data \cite[]{verschuur_seismic_2013}.
Although widely used, \gls{srme} is computationally intensive, depends on dense acquisition and high-quality near-offset traces, and often requires interpolation for optimal results \cite[]{verschuur_seismic_2013}.

Once a sufficiently accurate velocity model is available, alternative methods can be applied to \gls{nmo}-corrected \gls{cdp} gathers.
These methods exploit moveout differences between primaries and multiples.
Multiples typically remain unflattened following \gls{nmo} correction, which enables their separation from flat primaries.
The \gls{rt} is commonly used for this purpose: it maps \gls{cdp} gathers from the time–offset domain into the Radon domain, where events with different moveouts become separable before being transformed back \cite[]{hampson_inverse_1986, beylkin_discrete_1987}.
In this domain, a mute function is defined to isolate primaries from multiples based on moveout characteristics.
Parabolic \gls{rt} accomplishes this by representing the data as a sum of parabolic trajectories and reconstructing the input via a least-squares optimization in the Radon space \cite[]{hampson_inverse_1986}.
However, its resolution is limited, making it difficult to distinguish events with similar moveouts.

Several enhancements to \gls{rt} have been introduced to mitigate these limitations, including high-resolution formulations based on stochastic inversion in the time domain \cite[]{thorson_velocitystack_1985}, sparse inversion in the frequency domain \cite[]{sacchi_highresolution_1995}, and hybrid time–frequency approaches \cite[]{trad_latest_2003, lu_accelerated_2013}.
Nonetheless, these methods often require careful and problem-dependent hyperparameter tuning \cite[]{trad_latest_2003}.
Moreover, parabolic \gls{rt} assumes ideal parabolic event curvature, which is frequently violated in field data; similar challenges affect linear \cite[]{taner_long_1980, abbasi_attenuating_2013, verschuur_seismic_2013} and hyperbolic \gls{rt} variants \cite[]{foster_suppression_1992, verschuur_seismic_2013}.
A further practical drawback is that the mute function used to separate primaries and multiples is typically defined using a single reference \gls{cdp}, which may not generalize across a survey, making it necessary to pick mute zones on multiple \glspl{cdp} and interpolate between them.

Recently, supervised \gls{dl} has emerged as a promising alternative to conventional approaches, helping to overcome several limitations of traditional demultiple techniques.
For shot gathers, \gls{dl} models have been trained to perform the adaptive subtraction step of  \cite[]{zhang_deep_2021, li_feature_2020}, as well as to reconstruct near-offset traces required to enhance  performance \cite[]{qu_training_2021}.
Other work has shown that \glspl{nn} can approximate sparse-inversion–based primary estimation for suppressing surface-related multiples \cite[]{siahkoohi_surface-related_2019}.
\gls{dl}-based solutions have also been proposed for demultiple methods that rely on moveout discrimination in \gls{cdp} gathers.
For example, \glspl{cnn} have been trained to emulate the hyperbolic Radon transform and thereby separate primaries and multiples \cite[]{kaur_separating_2020}.
Several studies demonstrate \glspl{cnn} capable of predicting primary reflections from \gls{cdp} gathers that contain primaries mixed with residual multiples in post-migrated data \cite[]{nedorub_deep_2020, bugge_demonstrating_2021, fernandez_towards_2024}.
Additional architectures explored for post-migration demultiple include GAN-based models \cite[]{fernandez_deep_2023} and diffusion models \cite[]{durall_deep_2023}.

A major advantage of \gls{dl}-based demultiple approaches is that the model parameters are learned once during training, enabling a largely parameter-free and user-friendly inference stage.
This reduces computational cost and eliminates several labor-intensive steps—such as picking mute functions in the Radon domain or performing repeated parameter searches.
However, to the best of our knowledge, existing \gls{dl}-based approaches operate on individual \gls{cdp} gathers without incorporating information from neighboring \glspl{cdp} along the seismic line.
Processing \glspl{cdp} independently may lead to lateral inconsistencies in the demultiple results, i.e., removing certain events in one \gls{cdp} while retaining them in adjacent \glspl{cdp}, due to variations in velocity analysis and the resulting differences in moveout.

\begin{figure}[htbp]
\hspace{1cm}
\begin{subfigure}[t]{0.37\textwidth}
  \vspace{0pt}
  \centering
  \includegraphics[width=\textwidth]{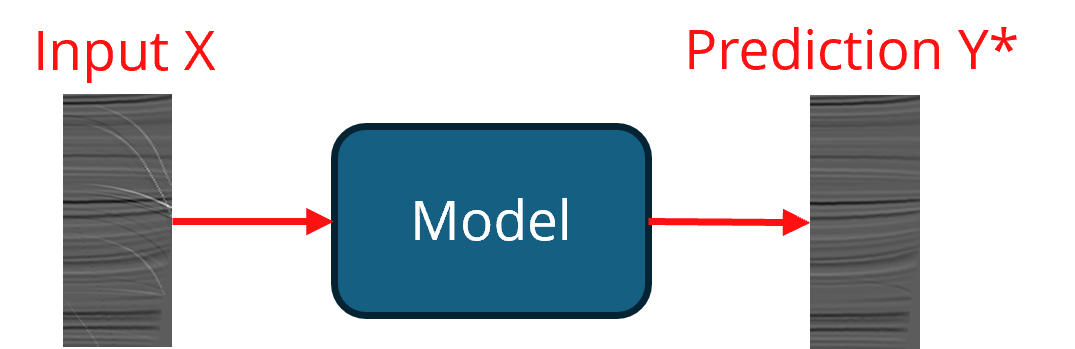}
  \vspace{2.47cm}
  \caption{Supervised Machine Learning}
  \label{fig:supervised}
\end{subfigure}
\hfill
\begin{subfigure}[t]{0.4\textwidth}
  \vspace{0pt}
  \centering
  \includegraphics[width=\textwidth]{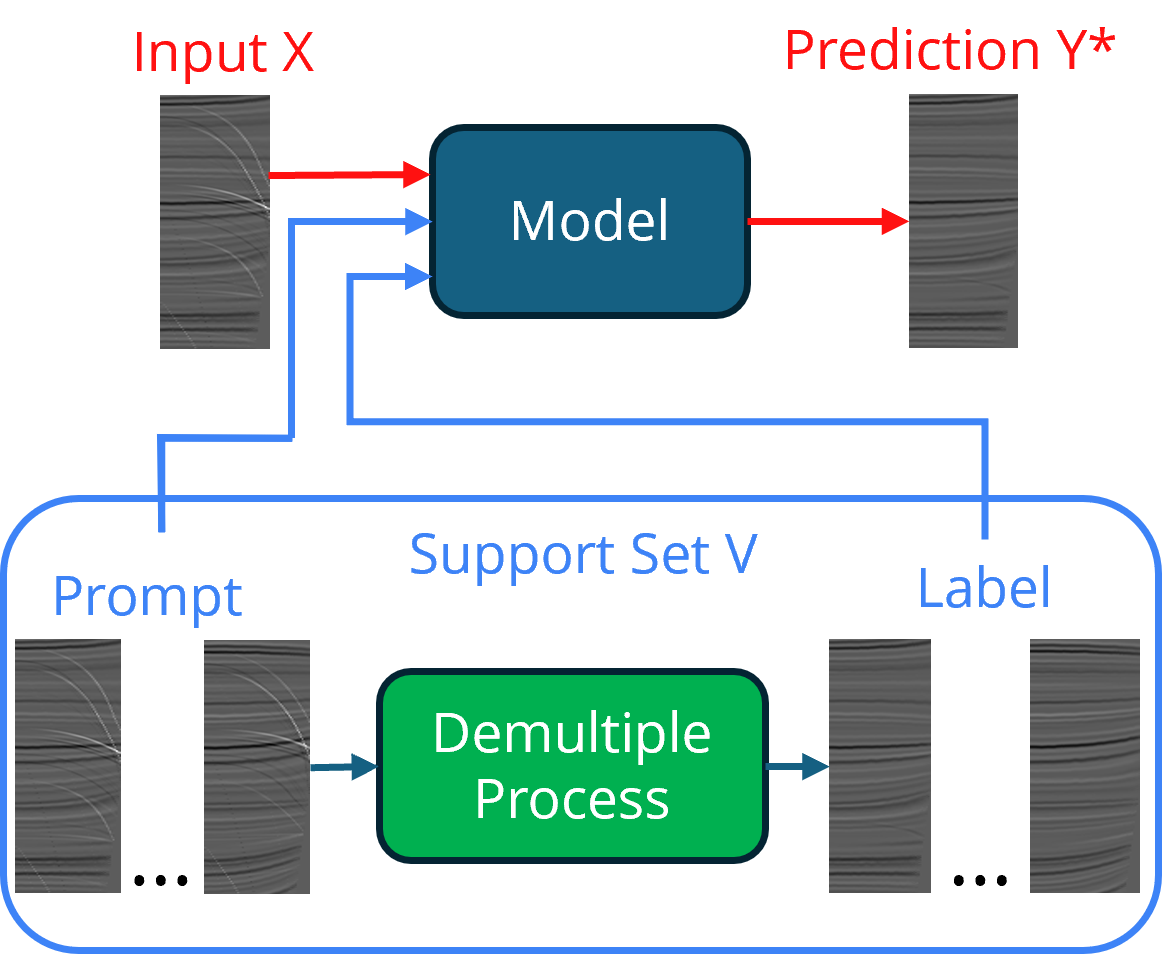}
  \caption{In-Context Learning}
  \label{fig:icl}
\end{subfigure}
\hspace{1cm}
\caption{Comparison of supervised learning and \glsxtrlong{icl}. Supervised Learning predicts $\mathbf{Y}^*$ based solely on $\mathbf{X}$. \Glsxtrlong{icl} predicts $\mathbf{Y}^*$ based on $\mathbf{X}$ and a support set $\mathcal{V}$ containing \glspl{cdp} from the same seismic line as $\mathbf{X}$ and labels obtained via an arbitrary demultiple process (e.g. Radon).}
\label{fig:supervised_visual_prompting_icl}
\end{figure}

One potential solution for enforcing lateral consistency is employing higher dimensional \glspl{nn}.
\cite{sansal_scaling_2025} introduced a three dimension \glspl{nn} for post-stack data.
Post-stack data are either two-dimensional sections (inline $\times$ time) or three dimensional cubes (inline $\times$ crossline $\times$ time), because the individual traces have already been summed across offsets (and often azimuths).
In contrast, pre-stack data retains the acquisition dimensions before stacking.
A pre-stack \gls{cdp} gather is typically organized as time $\times$ offset (and sometimes azimuth), and a survey-volume becomes inline $\times$ crossline $\times$  time $\times$ offset ($\times$ azimuth).
Processing this natively would require four- or even five-dimensional convolutions, with memory and compute that scale prohibitively with offset/azimuth sampling.
As a result, high-dimensional \glspl{nn} are often impractical for pre-stack processing at survey scale.

\Gls{icl} offers an alternative approach to achieve lateral consistency without resorting to computationally expensive higher-dimensional architectures.
\Gls{icl} originated in natural language processing, where \cite{brown_language_2020} demonstrated that \glspl{llm} could perform novel tasks without explicit fine-tuning by simply providing a few demonstration examples within the input prompt. Following its success in the language domain, \gls{icl} has been successfully adapted to \gls{cv} applications. Pioneering works have demonstrated its effectiveness across various vision tasks: \cite{wang_seggpt_2023} first introduced \gls{icl} for image segmentation, and then extended this concept to multiple dense prediction tasks in \cite[]{wang_images_2023}. \cite{butoi_universeg_2023} and \cite{rakic_tyche_2024} applied \gls{icl} to medical image segmentation. While these approaches primarily use context to learn new tasks, our approach leverages it to enforce lateral consistency across seismic gathers.

To understand how \gls{icl} achieves this, consider the fundamental difference between these paradigms.
Traditional supervised learning (Figure \ref{fig:supervised}) maps an input $\mathbf{X}$ directly to a prediction $\mathbf{Y}^*$ without auxiliary contextual information at inference.
In contrast, \gls{icl} (Figure \ref{fig:icl}) leverages a support set $\mathcal{V}$ that provides contextual examples.
For seismic processing, this mechanism enables the network to exploit spatial correlations among neighboring gathers, producing laterally consistent predictions along seismic lines.
Additionally, the support set facilitates adaptation to domain shifts between synthetic training data and field data.
Beyond consistency and adaptability, \gls{icl} introduces controllability, a desirable feature in seismic processing \cite{fernandez_towards_2024}.
By selecting appropriate examples for the support set $\mathcal{V}$, users can guide the network toward specific processing outcomes, providing interpretable control over \gls{dl} models that would otherwise operate as fixed black boxes.

In this paper, we introduce \model{}, a deep learning method for in-context seismic processing.
To our knowledge, this represents the first application of \gls{icl} to seismic data.
Unlike standard deep learning approaches that treat each seismic gather independently, our method processes gathers by leveraging neighboring ones as contextual information, thereby producing laterally consistent results.
We focus specifically on seismic multiple attenuation.
Experiments on both synthetic and field data demonstrate that our approach outperforms methods that process gathers independently, while also enabling conditioning on outputs from conventional processing workflows.
Our method thus provides a bridge between deep learning models and traditional seismic processing techniques, enhancing the overall quality and consistency of the results.
\section{Methodology}

We first establish the theoretical framework by contrasting conventional supervised learning with \gls{icl}.
We then detail our synthetic dataset generation, training procedure, and model architecture.

\subsection{Conventional Supervised Learning}

Conventional supervised learning maps inputs directly to outputs via a function $f_{\theta}$ parametrized by $\theta$ \eqref{eq:def_f}.
Let $\mathbf{X} \in \mathbb{R}^{M \times H \times W}$ denote a seismic line containing $M$ \glspl{cdp}, where $H$ and $W$ represent time (or depth) and offset (or angle) respectively.
For the  $m$-th \gls{cdp} $\notationXm$ the model predicts $\notationYm \in \mathbb{R}^{H \times W}$.

\begin{equation}
  f_{\theta}(\notationXm) =  \notationYm
  \label{eq:def_f}
\end{equation}

This approach treats each \gls{cdp} independently, disregarding spatial relationships between neighboring ones.
This independence introduces several limitations:
First, the approach fails to exploit the spatial continuity inherent in seismic data, where \glspl{cdp} exhibit smooth variations across neighboring positions.
Second, models can interpret similar events inconsistently across adjacent \glspl{cdp}, particularly when multiples intersect primaries with minimal moveout differences, resulting in lateral inconsistency.

\subsection{Supervised In-Context Learning}

To address these limitations,  \gls{icl}  conditions the input-output mapping on a support set $\notationVm$ enabling task-specific model behavior without retraining and increasing lateral consistency, as defined in \eqref{eq:def_f_icl}.

\begin{equation}
  f^{\text{ICL}}_{\theta} \left( \notationXm | \notationVm \right) = \notationYm
  \label{eq:def_f_icl}
\end{equation}

The support set $\notationVm = \{ \left( \mathbf{X}[s],f^{\mathcal{V}}(\mathbf{X}[s]) \right): TopS(sim(\mathbf{X}[m], \mathbf{X}[s])) > \tau \}$ consist of $S$ prompts $\mathbf{X}[s]$ and their corresponding labels $f^{\mathcal{V}}(\mathbf{X}[s])$.
The prompts $\mathbf{X}[s]$ come from \glspl{cdp} neighboring $\notationXm$ and the prompt-labels are generated by an arbitrary demultiple process $f^{\mathcal{V}}$. This enables line-specific adaptation, bridging the domain gap between synthetic training data and field data while incorporating prior knowledge through $f^{\mathcal{V}}$.


\subsection{Synthetic Dataset and Training}

To enable the model $f^{\text{ICL}}_{\theta}$ to effectively learn how to use the support set $\mathcal{V}$, we designed both a specialized training dataset and a corresponding training algorithm. For this purpose, it is not sufficient to rely on isolated \glspl{cdp} and their labels. Instead, we require spatially related \glspl{cdp} together with their associated primary and multiple events.

\begin{figure}[htbp]
  \centering
  \begin{subfigure}[b]{\textwidth}
    \centering
    \includegraphics[width=\textwidth]{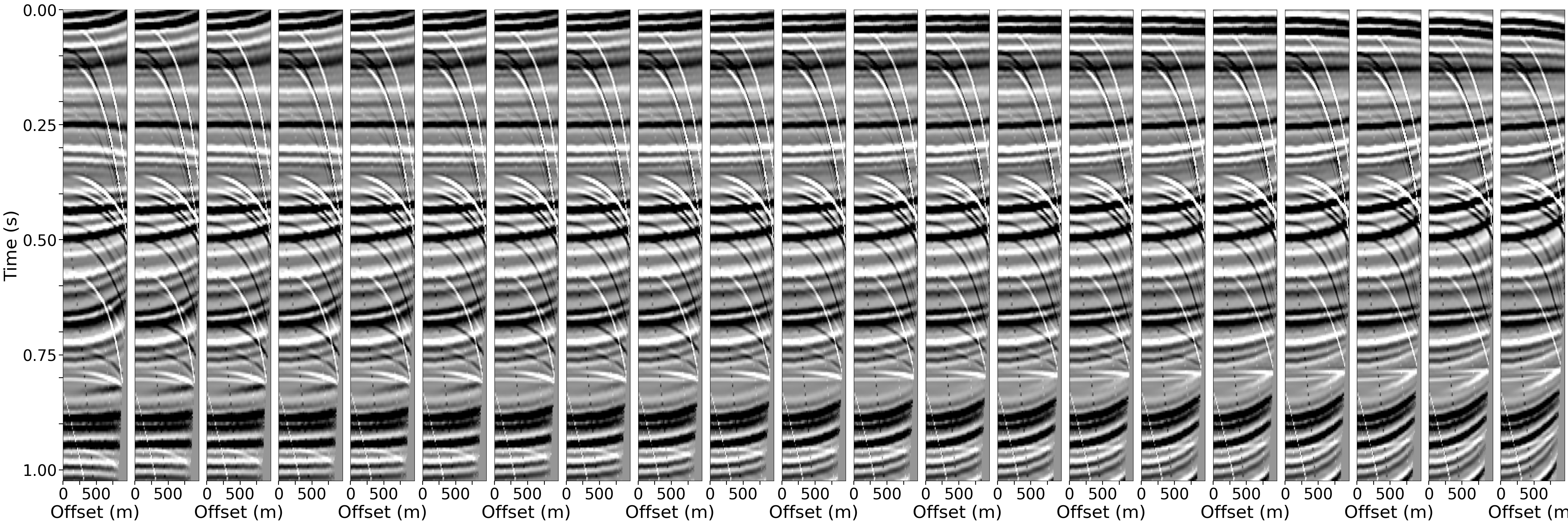}
    \caption{Gather}
  \end{subfigure}
  \hfill
  \begin{subfigure}[b]{\textwidth}
    \centering
    \includegraphics[width=\textwidth]{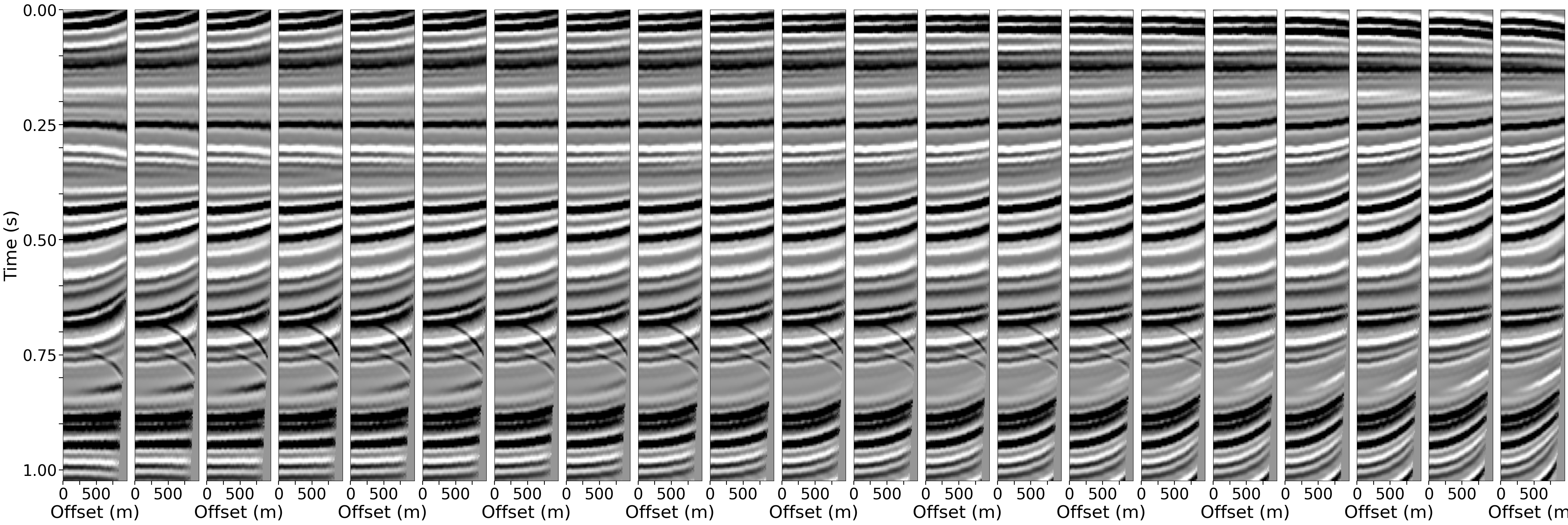}
    \caption{Label}
  \end{subfigure}
  \caption{Example of the spatially related gathers used during training.}
  \label{fig:synthetic_dataset}
\end{figure}

The synthetic seismic data were generated using convolutional modeling as described in \cite{fernandez_towards_2025}.
In that work, \glspl{cdp} containing both multiples and primaries were synthetically produced, while the corresponding labels contained only primaries, allowing the authors to train a model for seismic demultiple.
However, their training data treated each \gls{cdp} independently, whereas in real field acquisition and processing, \glspl{cdp} exhibit spatial continuity, with small variations across neighboring positions.
Ignoring this spatial relationship can lead to models that interpret similar events in neighboring \glspl{cdp} differently, especially in situations where multiples intersect primaries with little moveout difference.

In this work, we build on the study by \cite{fernandez_towards_2025} and generate spatially related \glspl{cdp}.
Spatially correlated primaries and multiples are created by introducing lateral variations in reflection coefficients across $M$ neighboring \gls{cdp} positions.
These coefficients are convolved with the source wavelet and then subjected to \gls{nmo} correction, yielding $M$ \glspl{cdp} that emulate a smoothly varying subsurface.
To increase realism, the \gls{nmo} correction is performed using spatially varying velocities, producing under-corrected or over-corrected events at different \gls{cdp} locations.
The same procedure is applied to generate multiple events but using moveouts characteristic of stronger curvature.

Each seismic gather is formed by combining the primary and multiple components, while the corresponding labels contain only the primaries. Every \gls{cdp} gather has a size of 64 traces and 256 time samples, and the final complete dataset consists of 15,000 synthetic seismic lines $\mathbf{X}$. Each seismic line consists of 21 spatially related \glspl{cdp} as presented in Figure \ref{fig:synthetic_dataset}.

During training, we randomly sample $S+1$ gather-label pairs from the $M$ spatially related gathers, where $S$ is the support set size.
One pair serves as input-output, while the remaining $S$ pairs form the support set $\mathcal{V}$.
After sampling, we apply data augmentations as detailed in Algorithm \ref{alg:training}.
First, we add random white noise to both gathers and labels, then normalize each pair by the gather's mean and standard deviation.
Additionally, we randomly replace a fixed percentage of training labels and prompt-labels with their corresponding inputs.
This introduces identity mapping examples that require the network to output the input unchanged.
This regularization technique prevents the model from solely learning the underlying demultiple transformation and instead encourages it to condition on the support set $\mathcal{V}$.

\begin{algorithm}
  \caption{\model{} training algorithm for one epoch with parameters: $\mathbf{X}^{\text{Train}}$ (training data consisting of $N$ seismic lines), $\mathbf{Y}^{\text{Train}}$ (corresponding label data), $N$ (number of seismic lines), $M$ (number of \glspl{cdp} in a line), $S$ (size of the support set $\mathcal{V}$), $\eta$ (learning rate), and $f^{\text{ICL}}_{\theta}$ (the model).}
  \label{alg:training}
  \begin{algorithmic}
    \Require $\mathbf{X}^{\text{Train}},\mathbf{Y}^{\text{Train}} \in \mathbb{R}^{N \times M \times H \times W}$
    \Ensure  $S \leq M $
    \For{$n = 0, \dots, N-1$}
    \LComment{Sample training data.}
    \State $i_0, i_2, \ldots, i_{S} \sim \text{DiscreteUniform}(0, M)$
    \Comment{Sample $S+1$ random indices.}
    \State $\mathbf{X} \gets \mathbf{X}^{\text{Train}}[n, i_0]$
    \Comment{First index $i_0$ is for the input $\mathbf{X}$ and output $\mathbf{Y}$.}
    \State $\mathbf{Y} \gets \mathbf{Y}^{\text{Train}}[n, i_0]$
    \For{$s = 1, \dots, S$}
    \Comment{Remaining indices are for the support set $\mathcal{V}$.}
    \State $\mathbf{V}[s-1, 0] \gets \mathbf{X}^{\text{Train}}[n, i_{s}]$
    \State $\mathbf{V}[s-1, 1] \gets \mathbf{Y}^{\text{Train}}[n, i_{s}]$
    \EndFor
    \LComment{Apply augmentations to the sampled data.}
    \State $\mathbf{X}, \mathbf{Y}, \mathbf{V} \gets \text{RandomWhiteNoise}(\mathbf{X}, \mathbf{Y}, \mathbf{V})$
    \Comment{Add random white noise to all the gathers as well as labels.}
    \State $\mathbf{X}, \mathbf{Y}, \mathbf{V} \gets \text{NormalizePerImage}(\mathbf{X}, \mathbf{Y}, \mathbf{V})$
    \Comment{Normalize each gather and label by the mean and standard deviation of the gather.}
    \State $\mathbf{X}, \mathbf{Y}, \mathbf{V} \gets \text{RandomlyReplaceLabel}(\mathbf{X}, \mathbf{Y}, \mathbf{V})$
    \Comment{For a specified percentage of steps replace the label with the input, as well as the prompt-labels with the prompts.}
    \LComment{Training.}
    \State $\mathbf{Y}^* \gets f^{\text{ICL}}_{\theta} \left(\mathbf{X}, \mathbf{V}\right)$ \Comment{Prediction.}
    \State $\ell \gets \mathcal{L}_1(\mathbf{Y}^*, \mathbf{Y})$
    \Comment{Compute loss.}
    \State $\theta \gets \theta - \eta \nabla_{\theta} \ell$
    \Comment{Update model weights.}
    \EndFor
  \end{algorithmic}
\end{algorithm}

\subsection{Model}

\begin{figure}[htbp]
  \centering
  \includegraphics[width=\textwidth]{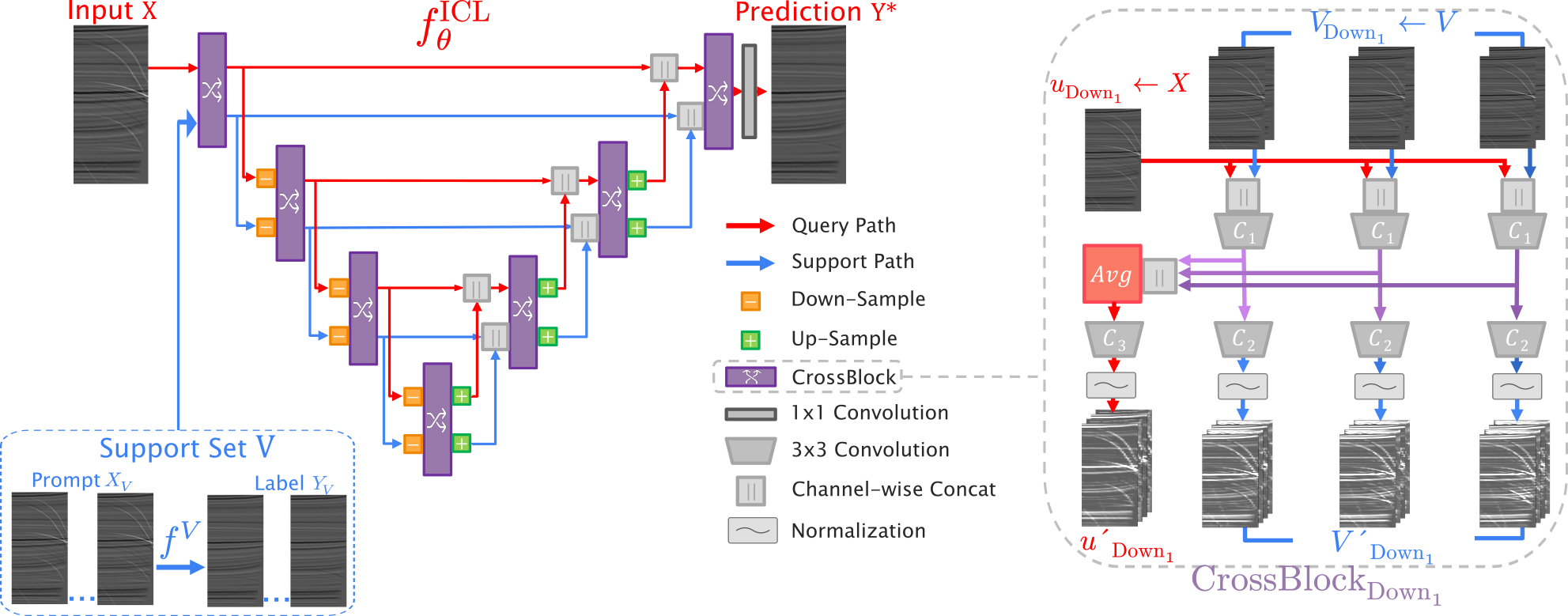}
  \caption{\model{} architecture based on a \unet{} with hierarchical features and skip-connections. Standard convolution blocks are replaced by CrossBlocks  \cite[]{butoi_universeg_2023} to enable interaction between the input and the support set. We modified the original CrossBlock design by adding normalization layers after the second convolutions. The model accepts a support set $\mathcal{V}$ as additional input, with $\mathcal{V}$ consisting of $S$ prompt gathers and their corresponding labels.}
  \label{fig:model}
\end{figure}

For the model $f^{\text{ICL}}_{\theta}$ we introduce \model{}, which is based on \universeg{} \cite[]{butoi_universeg_2023}, a medical image segmentation model that extends the \unet{} architecture \cite[]{ronneberger2015unetconvolutionalnetworksbiomedical} with task generalization capabilities.
While preserving the standard encoder-decoder structure with skip connections, \universeg{} replaces conventional convolutional blocks with CrossBlocks that perform cross-convolutions between query images and support examples, see Figure \ref{fig:model} and Equation \eqref{eq:crossblock}.
This modification enables inference with variable-sized support sets, allowing task specification at test time rather than through fixed training-time class mappings.

The CrossBlock conditions query features on a support set $\mathcal{V}$ through explicit cross-convolution operations.
Given a query feature map $\mathbf{u}$ and support feature maps $\mathcal{V}_{\text{feature}}$, the CrossBlock concatenates $\mathbf{u}$ with each feature map $\mathbf{V}_{\text{feature}}^s$ along the channel dimension, applies a shared convolution to all concatenated pairs \eqref{eq:crossblock_concat}, and averages the resulting interaction maps to update the query representation $\mathbf{u}'$ \eqref{eq:crossblock_update_feature}.
Additionally, each concatenated and convolved pair undergoes a second shared convolution to produce updated support representations $\mathcal{V}_{\text{feature}}'$ \eqref{eq:crossblock_update_support}.
The shared weights and averaging operation ensure permutation invariance and enable variable-sized support sets.

\begin{align}
  \text{CrossBlock}_{\theta_{C_1}, \theta_{C_2}, \theta_{C_3}} (\mathbf{u}, \mathcal{V}_{\text{feature}})
  &= (\mathbf{u}', \mathcal{V}_{\text{feature}}'),
  \quad \text{where}
  \label{eq:crossblock}
  \\
  \mathbf{u}' &=   \sigma \left( \text{Norm} \left( \text{Conv}_{\theta_{C_3}} \left(
        \frac{1}{S} \sum_{s=1}^{S} \mathbf{z}_s
  \right) \right) \right)
  \label{eq:crossblock_update_feature}
  \\
  \mathbf{z}_s &=  \text{Conv}_{\theta_{C_1}} \left(\mathbf{u} \| \mathbf{V}_{\text{feature}}^s \right)
  \label{eq:crossblock_concat}
  \\
  \mathbf{V}_{\text{feature}}^{\prime, s} &= \sigma \left( \text{Norm} \left( \text{Conv}_{\theta_{C_2}} \left(
        \mathbf{z}_s
  \right) \right) \right)
  \label{eq:crossblock_update_support}
\end{align}

We modified the original \universeg{} architecture by incorporating batch normalization after convolution operations in the CrossBlock \eqref{eq:crossblock}, which improved training stability in our experiments, see Appendix \ref{app:norm_layer}. Additionally, normalization techniques have been shown to accelerate training, improve gradient flow, and enhance model generalization in deep convolutional networks \cite[]{ioffe_batch_2015}. As activation function $\sigma$ we use a \gls{leakyrelu} \cite[]{maas_rectifier_2013}.

\section{Results}
\label{sec:results}

We evaluated our method through three approaches: quantitative analysis on our synthetic dataset's evaluation set ($15\%$ of the dataset), qualitative assessment of synthetic data results, and qualitative evaluation of field data performance.

\subsection{Baseline Models}
\label{sec:baseline_models}

For synthetic data evaluation, we trained a \unet{} baseline on the same dataset and for the same number of epochs as \model{}, with detailed training instructions available in the source code.
This baseline follows the architecture from \cite{fernandez_towards_2025} and processes each \gls{cdp} independently according to Equation \eqref{eq:def_f}.
Identical training configurations and training data ensure that performance differences arise solely from the architectural modifications introduced by \model{}, particularly the incorporation of support sets through CrossBlocks.

For field data evaluation, we employ a different \unet{} baseline from \cite{fernandez_towards_2025}, trained on 100,000 gathers.
The \unet{} trained on our smaller synthetic dataset exhibits poor field data generalization (see Appendix \ref{app:field_generalization}), whereas the larger-dataset model generalizes effectively.
This baseline difference illustrates a fundamental data efficiency limitation of conventional supervised learning that \model{} addresses through \gls{icl}.

\subsection{Quantitative Synthetic Results}

Although \model{} supports variable prompt numbers during training and inference, we used a fixed number of prompts due to computational constraints.
Variable tensor sizes require dynamic memory allocation and prevent efficient GPU parallelization.
This constraint raises the question of the optimal number of prompts to use during training.

\begin{figure}[htbp]
  \centering
  \includegraphics[width=\textwidth]{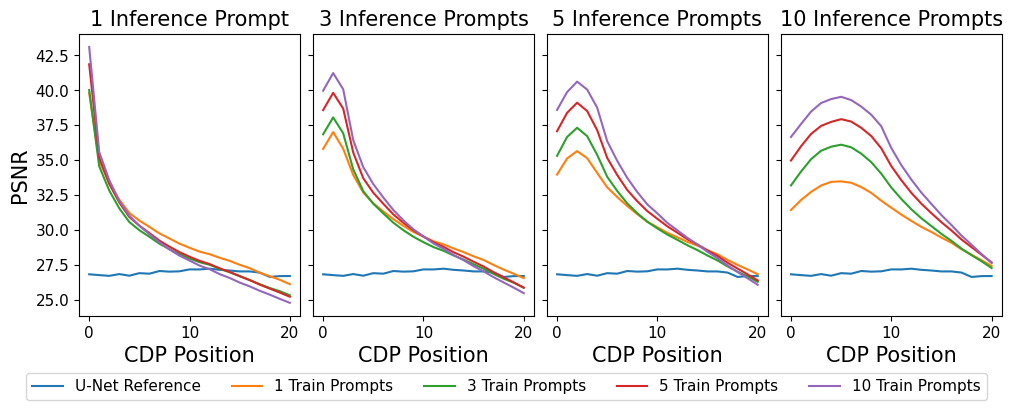}
  \caption{\Gls{psnr} versus \gls{cdp} position for \model{} models trained with varying numbers of prompts (1, 3, 5, 10) and a \unet{} baseline without prompting. Each subplot corresponds to a different number of inference prompts (1, 3, 5, 10), with the support set consisting of the first $S$ gathers and their corresponding labels. Increasing the number of inference prompts marginally reduces performance at early \gls{cdp} positions but improves the overall results. Models trained with more prompts exhibit enhanced performance at early \gls{cdp} positions across all inference conditions, while models trained with less prompts demonstrate superior performance at distant \gls{cdp} positions when using 1, 3, or 5 inference prompts.}
  \label{fig:synthetic_results_num_inference_prompts}
\end{figure}

\begin{figure}[htbp]
  \centering
  \includegraphics[width=\textwidth]{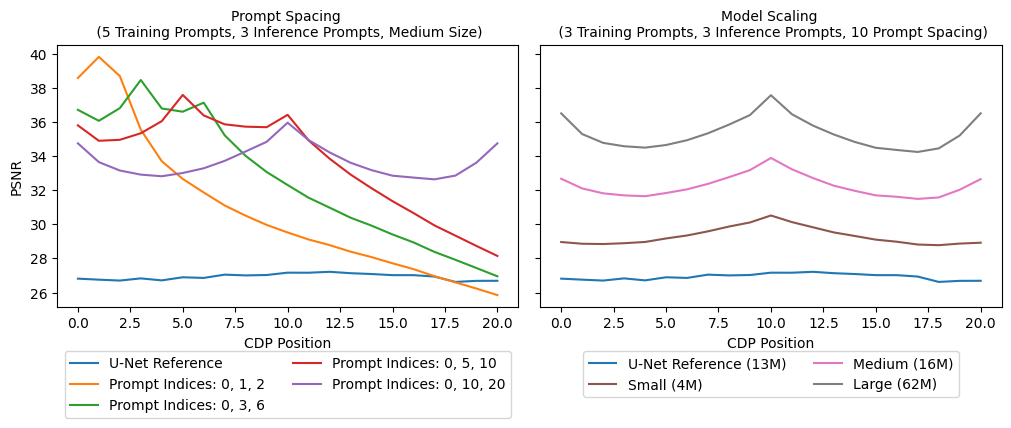}
  \caption{\Gls{psnr} of each \gls{cdp} against \gls{cdp} position. The left graph shows the effect of spacing the prompt differently on a medium-sized \model{} model trained with five prompts and evaluated with three inference prompts. For each prompt configuration the optimal results is at the center position of the prompts. Closer spaced prompts lead to better peak performance but also to significantly lower performance at the far \gls{cdp} positions. Using \gls{cdp} positions zero, ten and 20 seems to lead to consistent results for all \gls{cdp} positions. The right plot shows the scaling behavior of the \model{} model, demonstrating that larger \model{} architectures consistently outperform smaller variants, with even the smallest \model{} model surpassing the \unet{} baseline across all \gls{cdp} positions.}
  \label{fig:synthetic_results_prompt_spacing_model_scaling}
\end{figure}

Figure \ref{fig:synthetic_results_num_inference_prompts} shows \gls{psnr} versus \gls{cdp} position for \model{} models trained with 1, 3, 5, or 10 prompts, compared to our \unet{} baseline.
Each subplot represents different inference support set sizes (1, 3, 5, 10), with the support set consisting of the first $S$ gathers and their corresponding labels.
Increasing the number of inference prompts slightly degrades performance at early \gls{cdp} positions but improves overall results.
Models trained with more prompts show enhanced performance at early \gls{cdp} positions across all inference conditions, whereas models trained with fewer prompts achieve superior performance at distant \gls{cdp} positions when using one, three, or five inference prompts.

Beyond prompt quantity, prompt spacing significantly affects performance, as shown in the left subplot of Figure \ref{fig:synthetic_results_prompt_spacing_model_scaling}.
We evaluated a medium-sized \model{} model trained with five prompts using three inference prompts at different spacings.
Each configuration achieves optimal performance near the center of the prompt positions.
Closer prompt spacing improves peak performance at the \gls{cdp} positions where the prompts originate but degrades results at distant \gls{cdp} positions.
Wider spacing extends the range of high-quality results.
A spacing of ten \gls{cdp} positions between prompts provides consistent performance across all 21 \gls{cdp} positions.

The right subplot of Figure \ref{fig:synthetic_results_prompt_spacing_model_scaling}, demonstrates the effect of scaling the \model{} model.
All variants were trained with three prompts and evaluated using three prompts with a spacing of ten.
Larger models outperform smaller counterparts as expected, with even the 4-million parameter variant surpassing the \unet{} baseline across all \gls{cdp} positions.

Comparing the purple line (left subplot) with the pink line (right subplot) reveals that training with five prompts yields superior results to training with three prompts, despite identical model size and inference configuration.
This observation reinforces our findings from Figure \ref{fig:synthetic_results_num_inference_prompts}.

\subsection{Synthetic Examples}
\label{sec:synthetic_examples}

For qualitative evaluation, we selected the medium-sized \model{} trained with five prompts based on the quantitative analysis.
This configuration has a comparable parameter count to our \unet{} baseline and demonstrates superior performance on distant \glspl{cdp} relative to the ten-prompt variant.
During inference, we employed three prompts spaced every ten \gls{cdp} positions to balance prediction quality and annotation effort.
While ground-truth labels are available for synthetic data, practical deployment requires domain expert annotation to generate high-quality prompts.

Figure \ref{fig:synthetic_results_12850} demonstrates qualitative results of synthetic data: the first row displays spatially related gathers from our evaluation set, with corresponding ground truth labels in the second row. The third row presents predictions from our reference \unet{} model, while the fourth row shows the \model{} results.
These \model{} results exhibit enhanced spatial consistency across \gls{cdp}s, particularly evident between 0.25 and 0.5 seconds, and demonstrates superior performance for the near offsets around the 0.6-second event.

Appendix \ref{app:synthetic_results} presents additional synthetic examples that illustrate \model{} handling some challenging scenarios, like removing straight multiples that occur due to overcorrection of the primaries.

\begin{figure}[htbp]
  \centering
  \begin{subfigure}[b]{\textwidth}
    \centering
    \includegraphics[width=\textwidth]{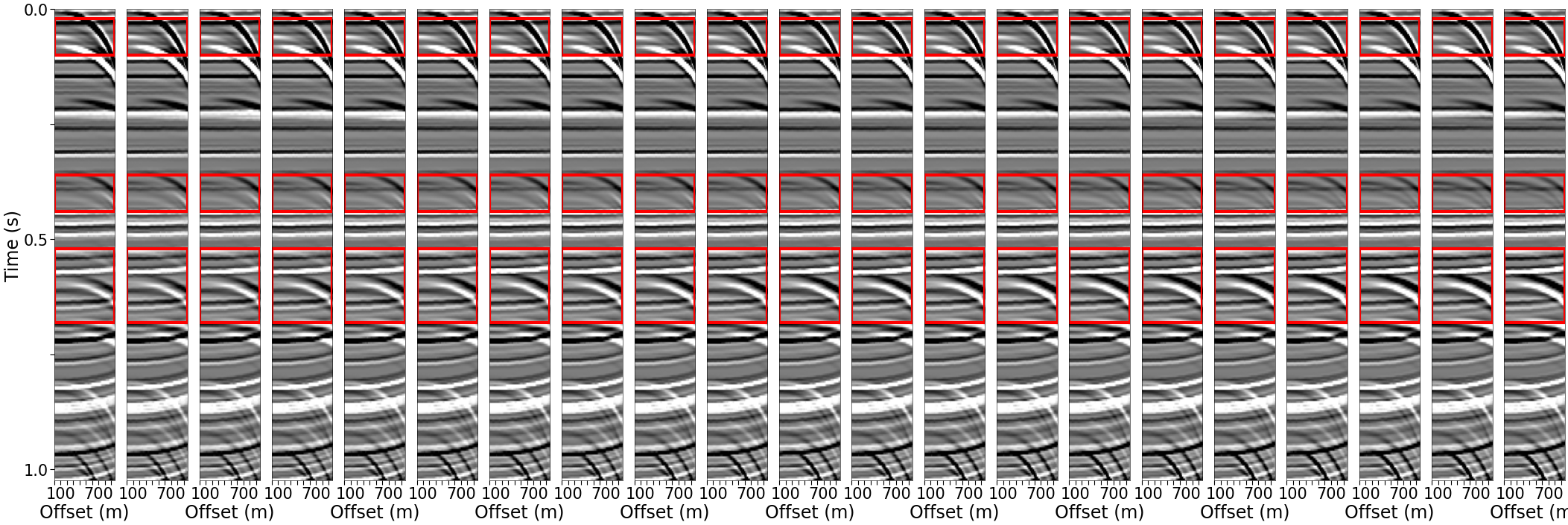}
    \caption{Gather}
  \end{subfigure}
\end{figure}
\begin{figure}[htbp]
  \ContinuedFloat
  \centering
  \begin{subfigure}[b]{\textwidth}
    \centering
    \includegraphics[width=\textwidth]{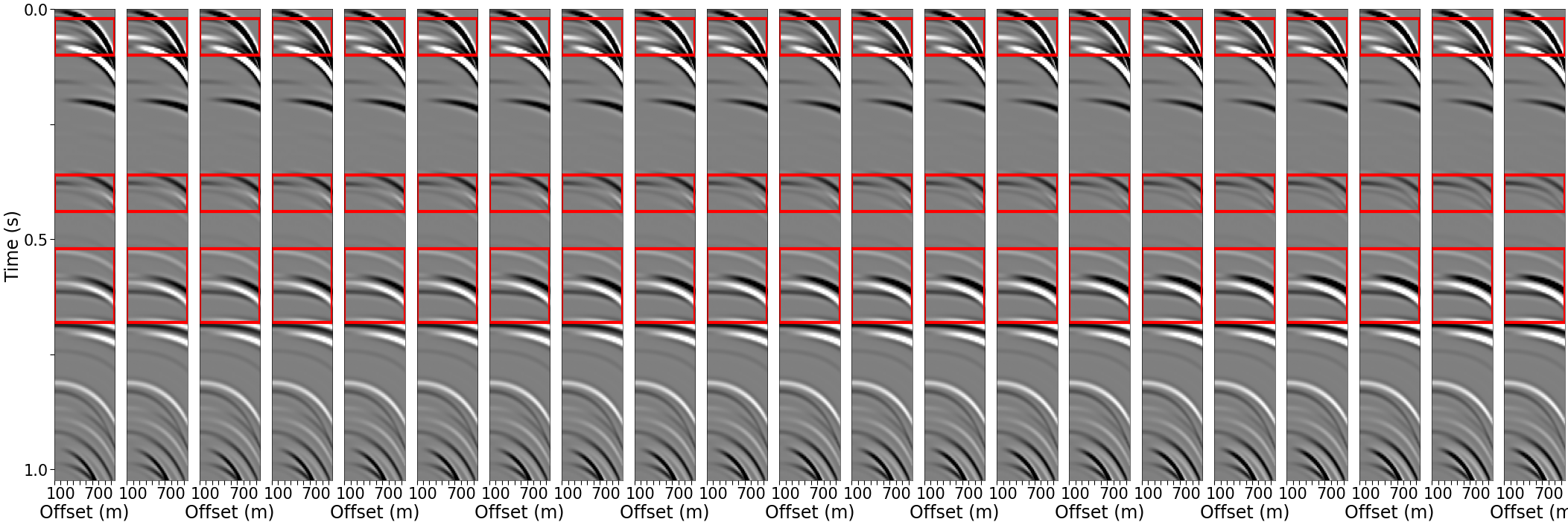}
    \caption{Multiples}
  \end{subfigure}
  \hfill
  \begin{subfigure}[b]{\textwidth}
    \centering
    \includegraphics[width=\textwidth]{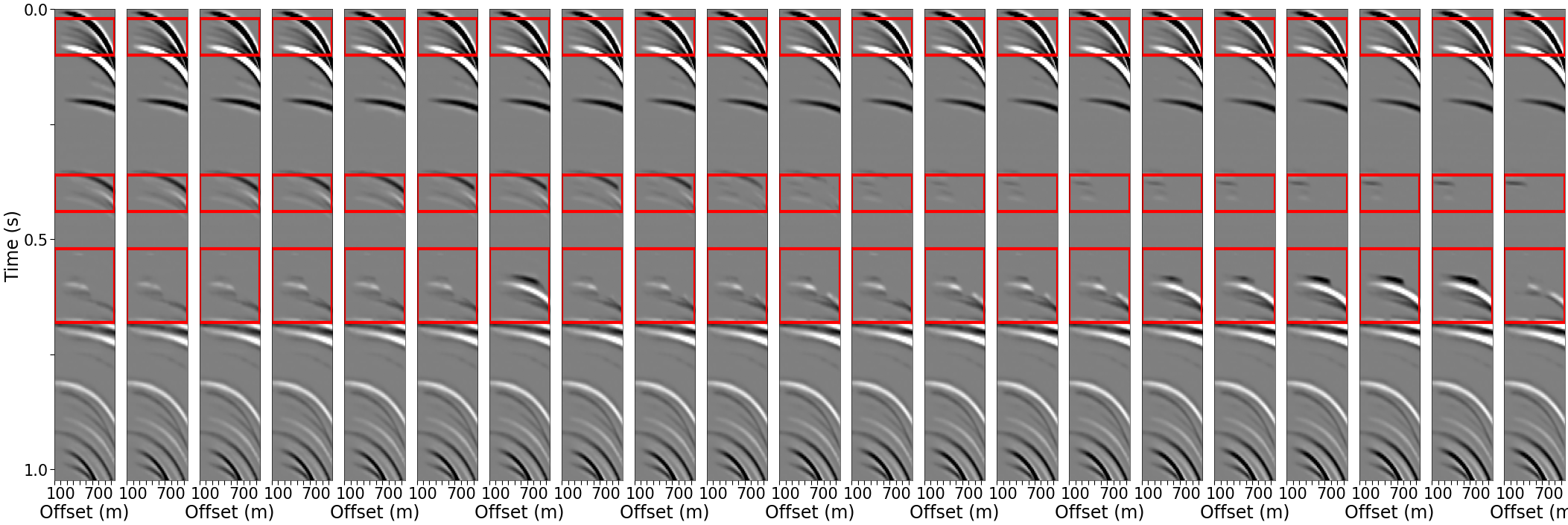}
    \caption{Predicted Multiples \unet{}}
  \end{subfigure}
  \hfill
  \begin{subfigure}[b]{\textwidth}
    \centering
    \includegraphics[width=\textwidth]{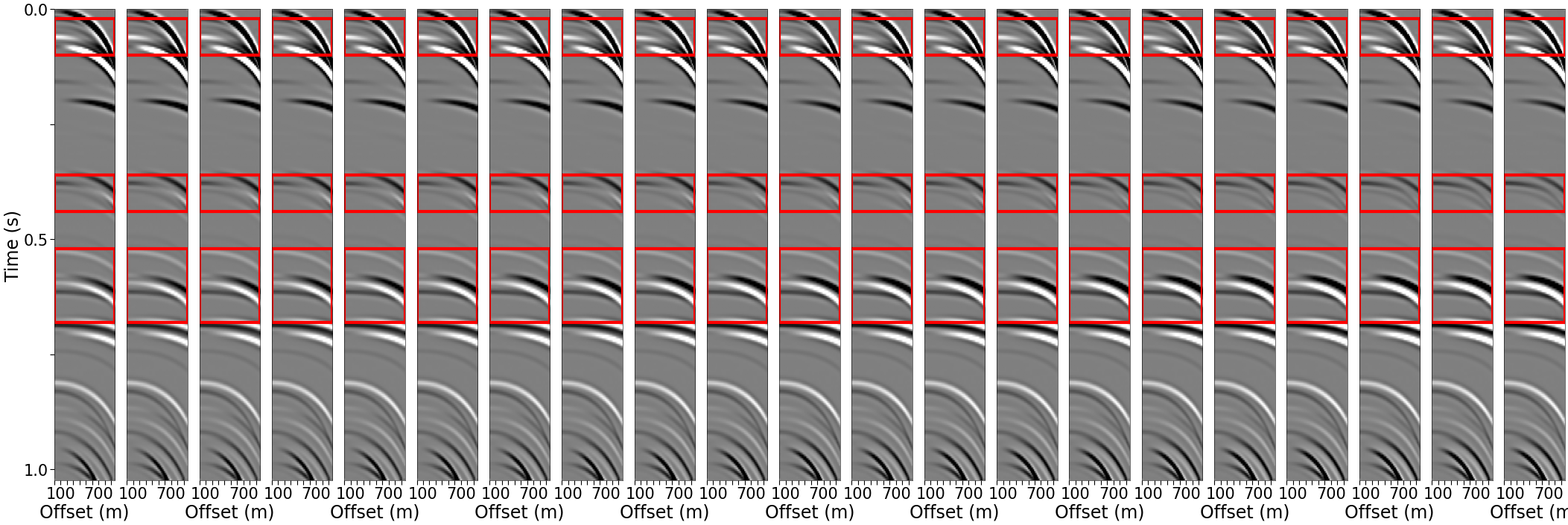}
    \caption{Predicted Multiples \model{}}
  \end{subfigure}
  \caption{Synthetic data results comparing our \unet{} reference, and our \model{} model to the ground truth. The latter shows significantly improved consistency across the \gls{cdp}s (visible for the events between 0.25 and 0.5 seconds), as well as notably improved behavior for the near offsets.}
  \label{fig:synthetic_results_12850}
\end{figure}

\subsection{Field Data Examples}

We now evaluate field data performance following the synthetic data analysis.
The dataset comprises post-migration \glspl{cdp} containing residual multiples.
Each \gls{cdp} consists of 63 traces spanning \SI{5}{\second}, sampled at \SI{4}{\milli\second} intervals.
The crossline spacing is \SI{25}{\meter}.
The data exhibit weak-amplitude, high-frequency parabolic multiples and linear noise.

Figures \ref{fig:field_results_consistency} and \ref{fig:field_results_near_offsets} present the same \gls{cdp}s at different time intervals, each highlighting distinct  characteristics. Both figures follow identical layouts: complete gathers with highlighted time slices (first row), multiples of a high resolution Radon \cite[]{sacchi_highresolution_1995} (second row), multiples of a baseline \unet{} \cite[]{fernandez_towards_2025} (third row), multiples of our \model{} model (fourth row), and primaries of our \model{} model(fifth row).
The \unet{} results correspond to the model trained with 100,000 gathers as discussed in Section \ref{sec:baseline_models}.
The \model{} results employ the same model as in Section \ref{sec:synthetic_examples}, with one modification: we apply three sequential prompts (\glspl{cdp} 1060, 1061, 1062) during inference instead of the sparse prompting strategy used for synthetic data.
This adjustment is necessary due to the lateral inconsistencies in the Radon results, which are used as prompts.

Figure \ref{fig:field_results_consistency} examines the 1-2 second interval, revealing consistency issues in both the traditional Radon and \unet{} methods. The event at 1.3 seconds is inconsistently removed across \gls{cdp}s by both methods, while \model{} maintains consistent removal across all \gls{cdp}s.

Figure \ref{fig:field_results_near_offsets} focuses on the 3.5-4.5 second interval, demonstrating the issues of the \unet{} model at near offsets. Both traditional Radon and \model{} achieve complete event removal, notably outperforming the \unet{} in this regard.
However, this figure also underscores the importance of high-quality prompts.
While \model{} exhibits better lateral continuity than both Radon and \unet{}, it closely follows the Radon prompts by design.
Consequently, it inherits the excessive removal present in our Radon results, potentially removing events that are not are not necessarily multiples, given that the Radon demultiple had been executed with a quite aggressive parameter set.

\begin{figure}[htbp]
  \centering
  \includegraphics[width=\textwidth]{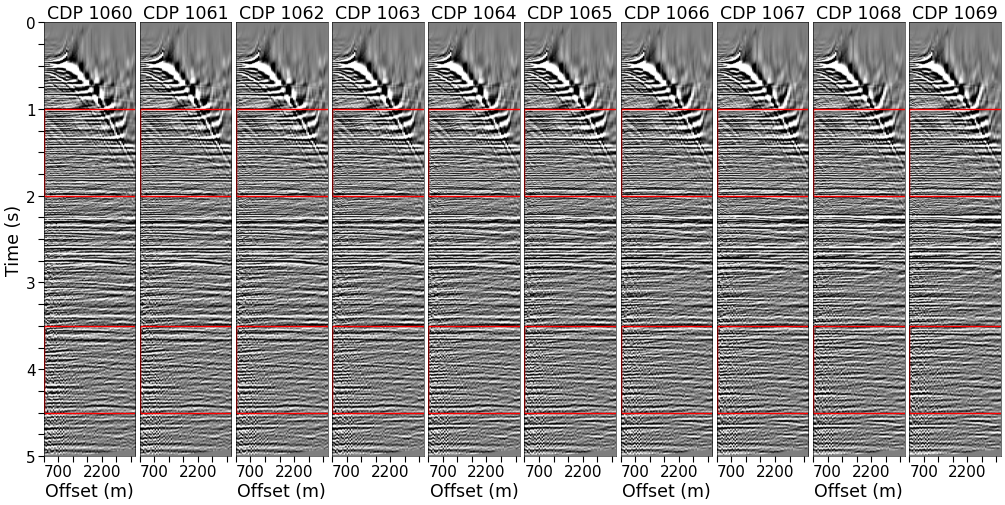}
  \caption{Post-migration data of a North Sea field with two areas highlighted that we want to investigate in the next two figures.}
  \label{fig:field_data}
\end{figure}

\begin{figure}[htbp]
  \centering
  \begin{subfigure}[b]{\textwidth}
    \centering
    \includegraphics[width=\textwidth]{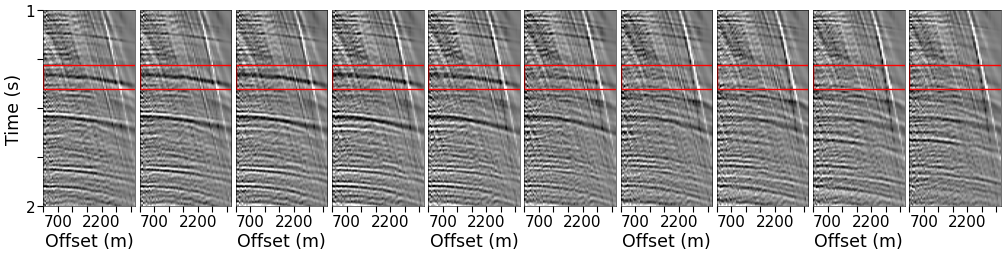}
    \caption{Multiples Radon}
  \end{subfigure}
  \hfill
  \begin{subfigure}[b]{\textwidth}
    \centering
    \includegraphics[width=\textwidth]{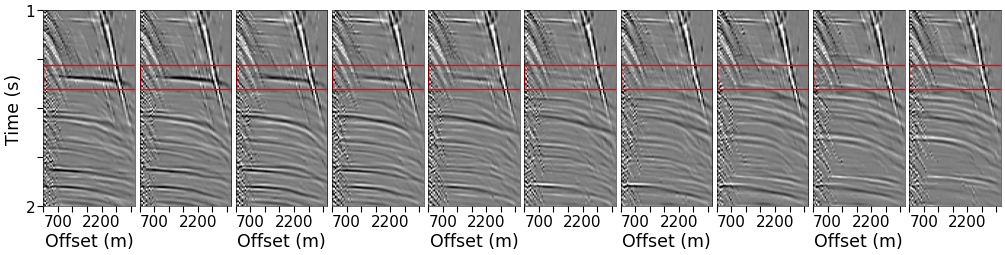}
    \caption{Multiples \unet{}}
  \end{subfigure}
  \hfill
  \begin{subfigure}[b]{\textwidth}
    \centering
    \includegraphics[width=\textwidth]{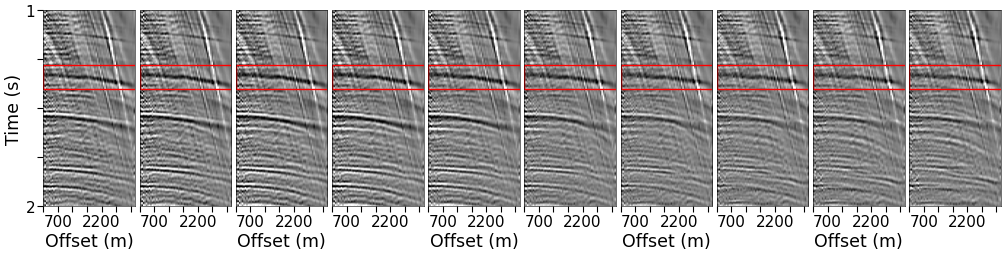}
    \caption{Multiples \model}
  \end{subfigure}
  \hfill
  \begin{subfigure}[b]{\textwidth}
    \centering
    \includegraphics[width=\textwidth]{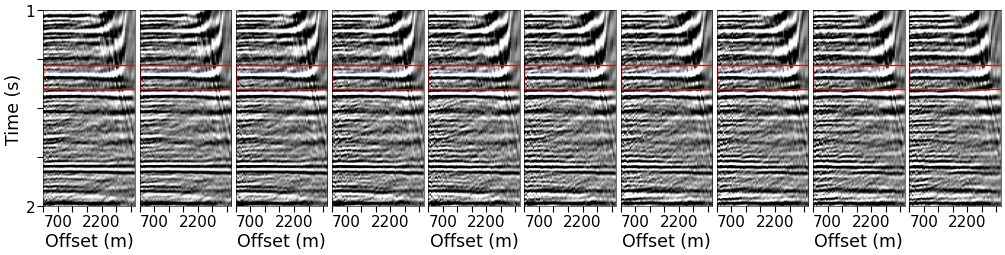}
    \caption{Primaries \model}
  \end{subfigure}
  \caption{Results for the upper highlighted area of the field data shown in Figure \ref{fig:field_data} comparing traditional Radon demultile, a \unet{} baseline \cite[]{fernandez_towards_2025}, and our \model{} model. The latter shows notably improved consistency across the \gls{cdp}s compared to both the traditional Radon demultiple results and the \unet{} baseline.}
  \label{fig:field_results_consistency}
\end{figure}

\begin{figure}[htbp]
  \centering
  \begin{subfigure}[b]{\textwidth}
    \centering
    \includegraphics[width=\textwidth]{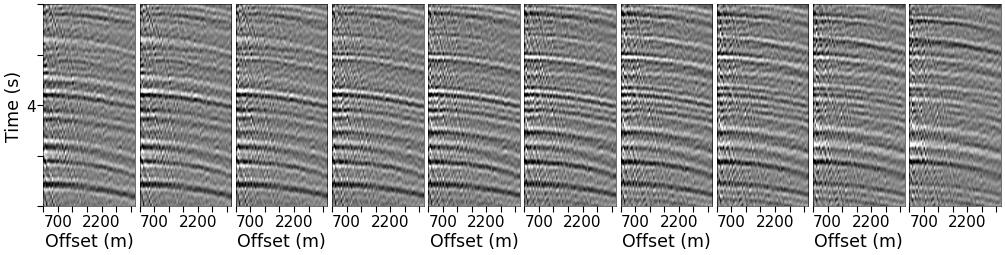}
    \caption{Multiples Radon}
  \end{subfigure}
  \hfill
  \begin{subfigure}[b]{\textwidth}
    \centering
    \includegraphics[width=\textwidth]{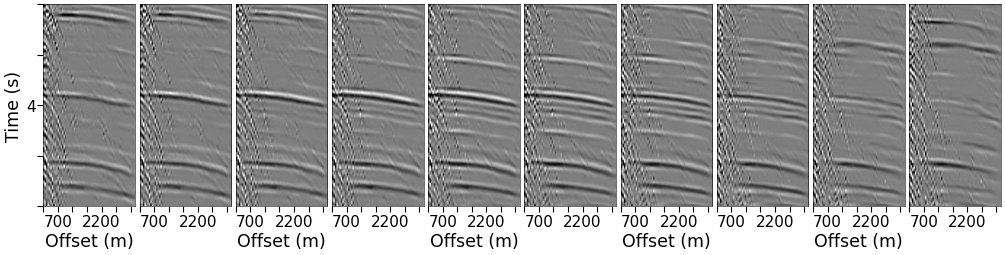}
    \caption{Multiples \unet{}}
  \end{subfigure}
  \hfill
  \begin{subfigure}[b]{\textwidth}
    \centering
    \includegraphics[width=\textwidth]{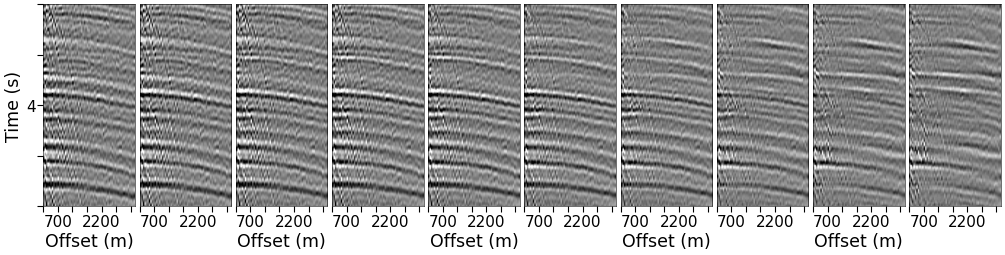}
    \caption{Multiples \model}
  \end{subfigure}
  \hfill
  \begin{subfigure}[b]{\textwidth}
    \centering
    \includegraphics[width=\textwidth]{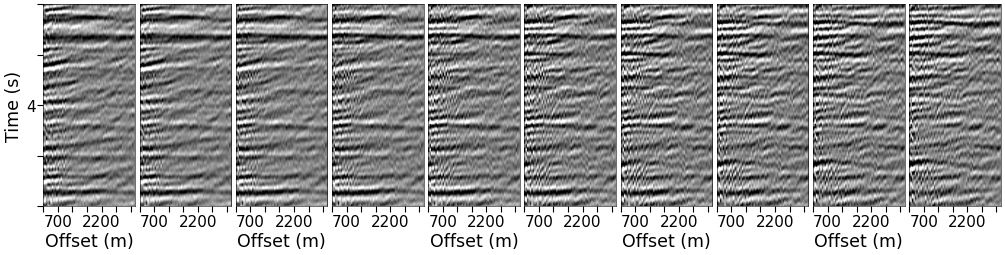}
    \caption{Primaries \model}
  \end{subfigure}
  \caption{Results for the lower highlighted area of the field data shown in Figure \ref{fig:field_data} comparing traditional Radon demultiple, a \unet{} baseline \cite[]{fernandez_towards_2025}, and our \model{} model. The latter shows notably improved results for the near offsets compared to the \unet{} baseline.}
  \label{fig:field_results_near_offsets}
\end{figure}

\section{Discussion}

Our results demonstrate that \gls{icl} improves seismic demultiple processing through two key improvements: increased consistency across \glspl{cdp} and more complete event removal at near offsets.
Building on these findings, this section examines the generalization capabilities and user-control mechanisms of \model{} and discusses strategies for prompt selection and training modifications to improve field inference. 
Finally, we identify additional seismic applications that could benefit from this approach.

\subsection{Generalization with Limited Training Data}

The \gls{icl}  approach exhibits superior data efficiency compared to conventional supervised learning.
While standard \unet{} models require approximately 100,000 gathers for adequate generalization \cite[]{fernandez_towards_2025}, our prompt-based model achieves comparable performance on field data with only 10,500 training gathers, as shown in Appendix \ref{app:field_generalization}.
This efficiency stems from the model's ability to leverage contextual information from the support set during inference.

\subsection{User Control Through Prompting}

\model{} provides direct user control over network outputs while bridging conventional and deep learning methodologies.
Traditional Radon demultiple results can serve directly as prompts, integrating established processing techniques with neural networks.
Alternatively, prompts can be constructed from depth-dependent stitching of existing deep learning outputs \cite[]{durall_deep_2023, fernandez_towards_2025}, or from hybrid combinations of Radon and deep learning results.
This flexibility allows users to guide predictions based on domain expertise and quality requirements.

\subsection{Prompt Selection Strategies}

\begin{figure}[htbp]
  \thisfloatsetup{capposition=beside,capbesideposition={right,top}}
  \fcapside[\FBwidth]{\caption{\Gls{mse} and standard deviation of ten identically trained models versus \gls{cdp} position. The models were trained with five prompts and three prompts were used during inference, with the inference prompts consisting of the first three prompt-prompt label pairs. The variance between the model predictions is highly correlated to the quality of the predictions and the \gls{cdp} position.}\label{fig:synthetic_variance}}{\includegraphics[width=0.5\textwidth]{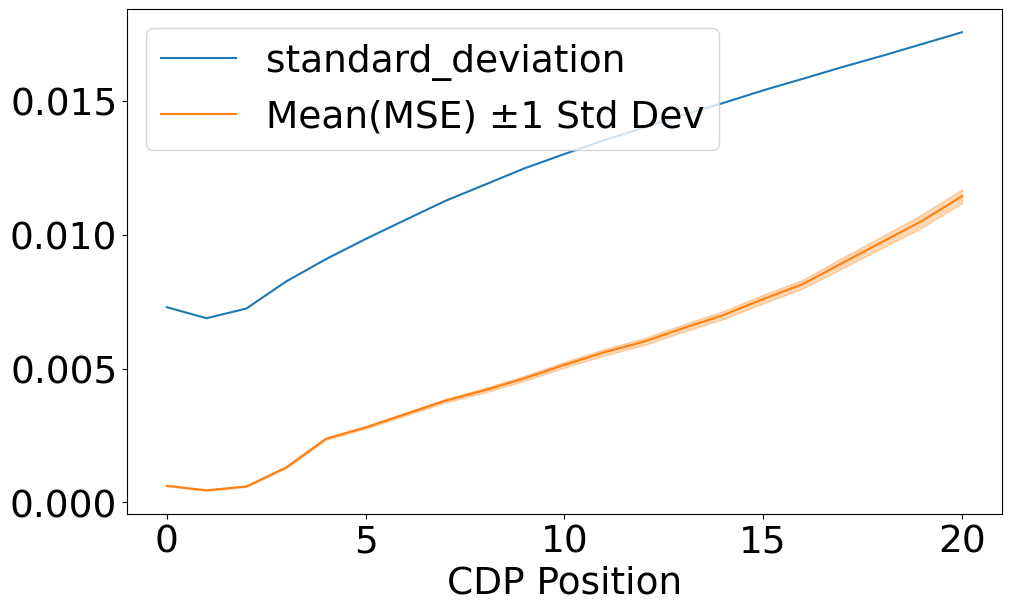}}
\end{figure}

For large seismic lines and volumes, prompt selection strategies require further investigation.
Adaptive re-prompting based on prediction variance from an ensemble of \model{} models shows promise, given the observed correlation between variance and result quality (Figure \ref{fig:synthetic_variance}).
Another strategy sequential re-prompting, where previously processed \gls{cdp}s guide subsequent predictions, suffers from substantial error propagation in preliminary tests.
Future work should investigate this on longer seismic lines or even seismic volumes.

\subsection{Perspectives}

Computational constraints restrict training to fixed-size support sets, as variable tensor sizes require dynamic memory allocation and prevent efficient GPU parallelization.
However, since variable-sized support sets are desired and used during inference, incorporating them during training may enhance inference performance.

Performance could also be enhanced using datasets with multiple labels, as in \cite{fernandez_towards_2024}, to increase the model's reliance on the support set.
Currently, we achieve this by randomly substituting labels and prompt-labels with their corresponding gathers for a fixed percentage of iterations.

Beyond demultiple processing, the \gls{icl} framework shows potential for broader seismic applications.
The methodology's advantages, user-guided predictions and enhanced spatial consistency through contextual information, could extend to other processing tasks such as alignment and destretch.
Similarly, seismic interpretation workflows, including fault detection, salt-body delineation, and horizon picking, could benefit from both user control and improved consistency by leveraging contextual information along seismic lines or across volumes.

\section{Conclusion}

To our knowledge, this work presents the first application of \gls{icl} to seismic processing.
We introduced \model{} an adaptation of \universeg{}, a medical image segmentation model, to address limitations in spatial consistency and processing flexibility in current deep learning approaches for seismic data processing.
Our experiments demonstrate that \gls{icl} provides significant improvements over conventional \unet{} architectures in two key areas: improved spatial consistency across gathers and better near-offset performance.
Furthermore, it requires substantially less training data than conventional \unet{}.

The flexible prompting strategy enables integration of traditional processing methods with deep learning capabilities.
Both Radon demultiple results and depth-dependent stitches of existing deep learning outputs can serve as effective prompts.
Quantitative analysis reveals improvements over the reference \unet{} and demonstrates a strong correlation between the quality of the results and the lateral distance from the input to the prompts.
Additionally, field data validation confirms improved consistency compared to both traditional Radon demultiple and baseline \unet{} methods.
The methodology's success in seismic demultiple processing suggests potential applicability to other seismic processing tasks requiring spatial consistency and user control.

\section{Acknowledgments}

The authors would like to acknowledge the members of the Fraunhofer ITWM DLSeis Consortium (http://dlseis.org) for their financial support.
We appreciate Equinor ASA, V\r{a}r Energy ASA, Petoro AS and ConocoPhillips Skandinavia AS for granting us permission to utilize their field data.

\section{Data and Materials availability}

Part of the data used in this research can be obtained by contacting the authors upon reasonable request.
The source code is available at: \url{https://codeberg.org/fuchsfa/in-context-learning-seismic}.
In addition to the source code, the experimental metrics and settings are also saved in the Git repository and accessible via DVC.

\newpage
\bibliographystyle{unsrtnat}
\bibliography{ICL_Paper}


\newpage
\appendix

\section{Norm Layer}
\label{app:norm_layer}

\begin{figure}[htbp]
  \thisfloatsetup{capposition=beside,capbesideposition={right,top}}
  \fcapside[\FBwidth]{\caption{Comparison of five training runs with and without the BatchNorm layer we added to \model{}. The plot shows the L1 loss averaged over the five training runs with a confidence area of one standard deviation. The training runs with the Normlayer performed significantly more stable and led to a smaller loss in the end. The line showing the trainings without Normlayers is not fully visible due to drastic divergence in some of training runs.}\label{fig:norm_layer}}{\includegraphics[width=0.7\textwidth]{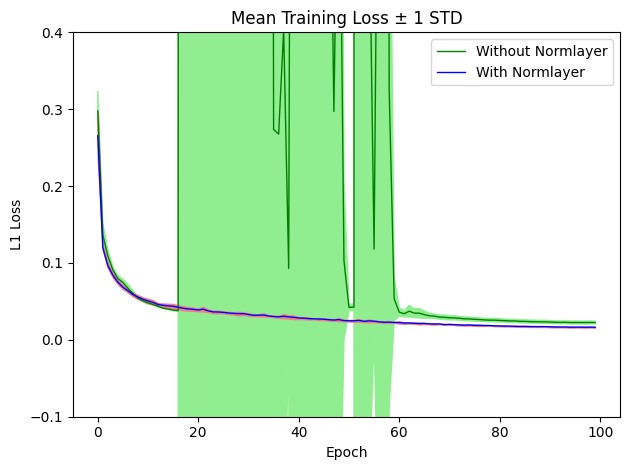}}
\end{figure}

Figure \ref{fig:norm_layer} reports the mean L1 loss across five runs, with shaded bands indicating a confidence area of one standard deviation, for models trained with and without the BatchNorm layer we added to \model{}. All experiments used the small \model{} with batch size 64, optimized the L1 loss with AdamW (learning rate 0.001, weight decay 0.01), applied a OneCycle learning rate schedule, and clipped gradients to a maximum norm of 1.

The blue line with the pink band denotes the trainings with BatchNorm and the green line with the light green band denotes the trainings without BatchNorm. The trainings with BatchNorm yield lower variance and greater training stability. In contrast, several runs without BatchNorm diverged strongly, rendering portions of the average loss curve not visible.

\section{Synthetic Results}
\label{app:synthetic_results}

This section presents additional examples of synthetic data results to demonstrate the performance differences between \unet{} and \model{}. Each figure consists of: (a) the input gather, (b) the ground truth multiples, (c) multiples predicted by the \unet{} baseline, and (d) multiples predicted by our \model{}.

\begin{figure}[hb]
  \centering
  \begin{subfigure}[b]{\textwidth}
    \centering
    \includegraphics[width=\textwidth]{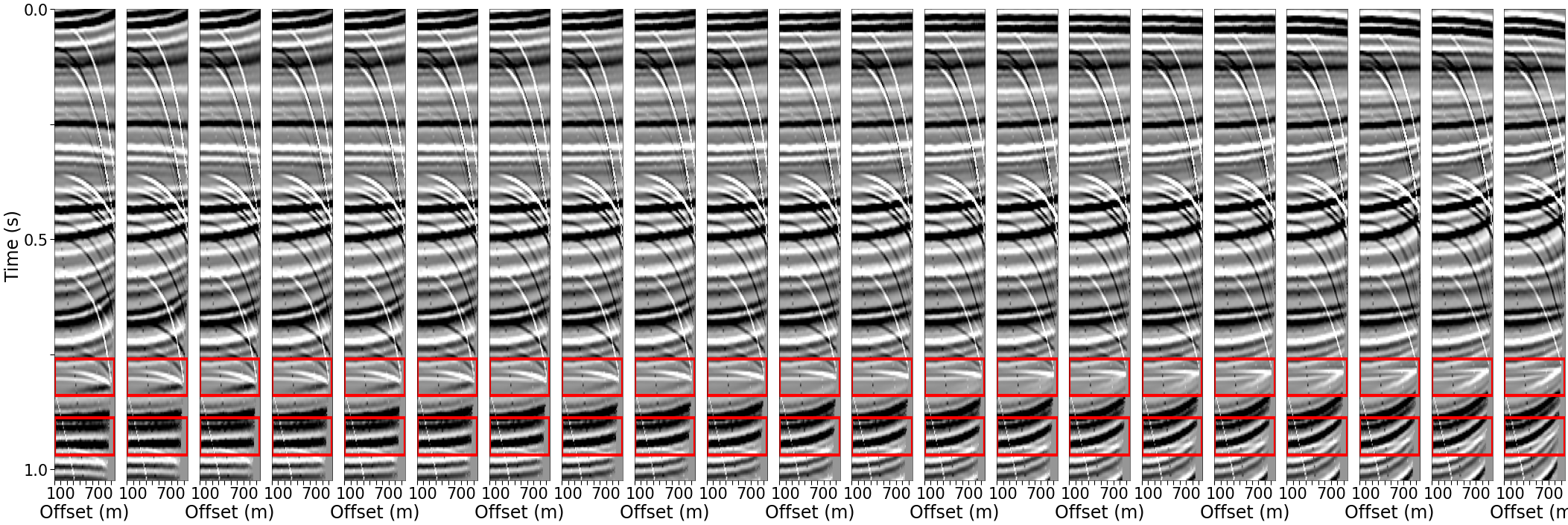}
    \caption{Gather}
  \end{subfigure}
  \hfill
  \begin{subfigure}[b]{\textwidth}
    \centering
    \includegraphics[width=\textwidth]{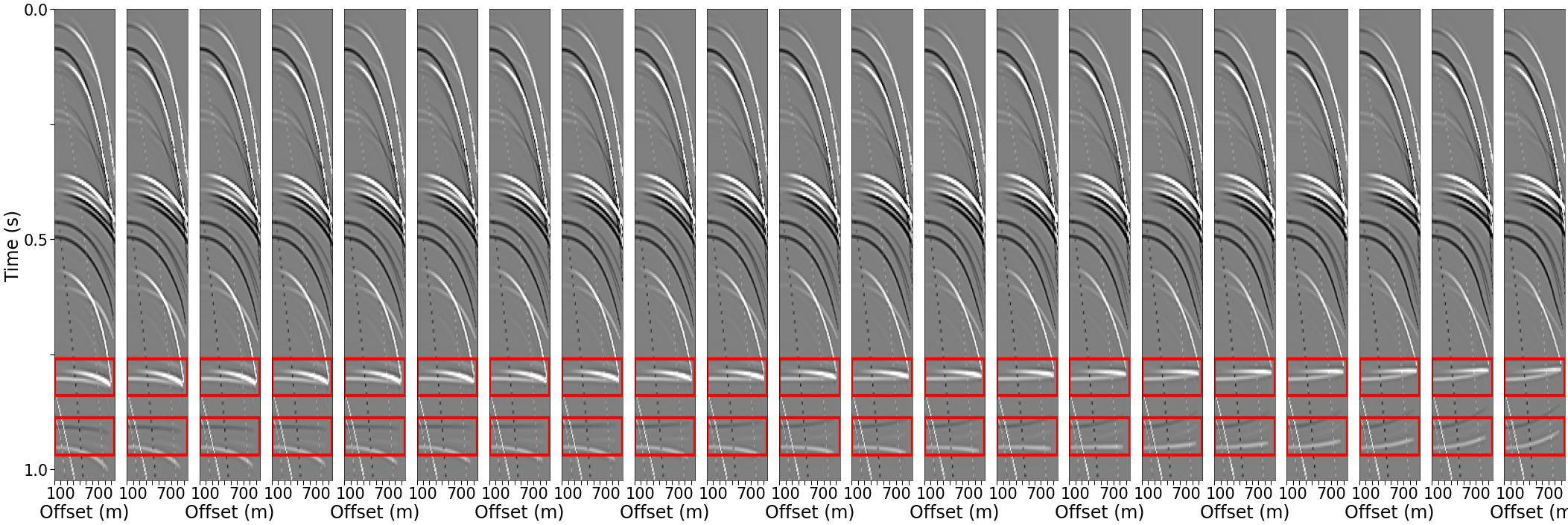}
    \caption{Multiples}
  \end{subfigure}
  \hfill
  \begin{subfigure}[b]{\textwidth}
    \centering
    \includegraphics[width=\textwidth]{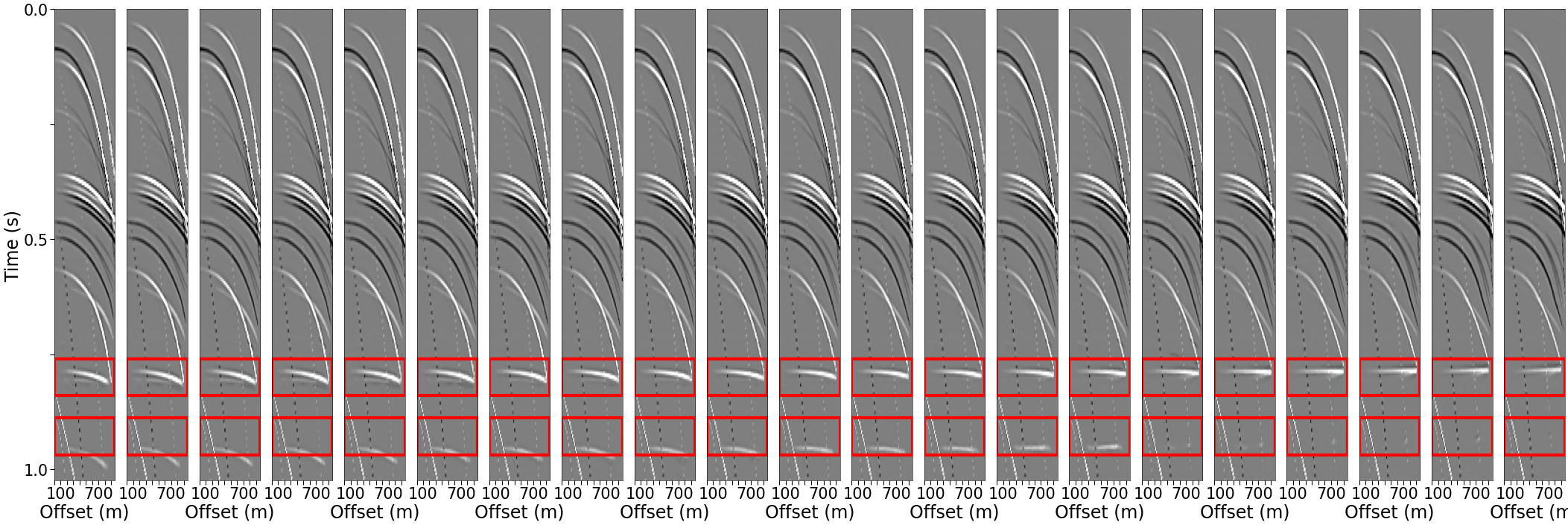}
    \caption{Predicted Multiples \unet{}}
  \end{subfigure}
  \hfill
  \begin{subfigure}[b]{\textwidth}
    \centering
    \includegraphics[width=\textwidth]{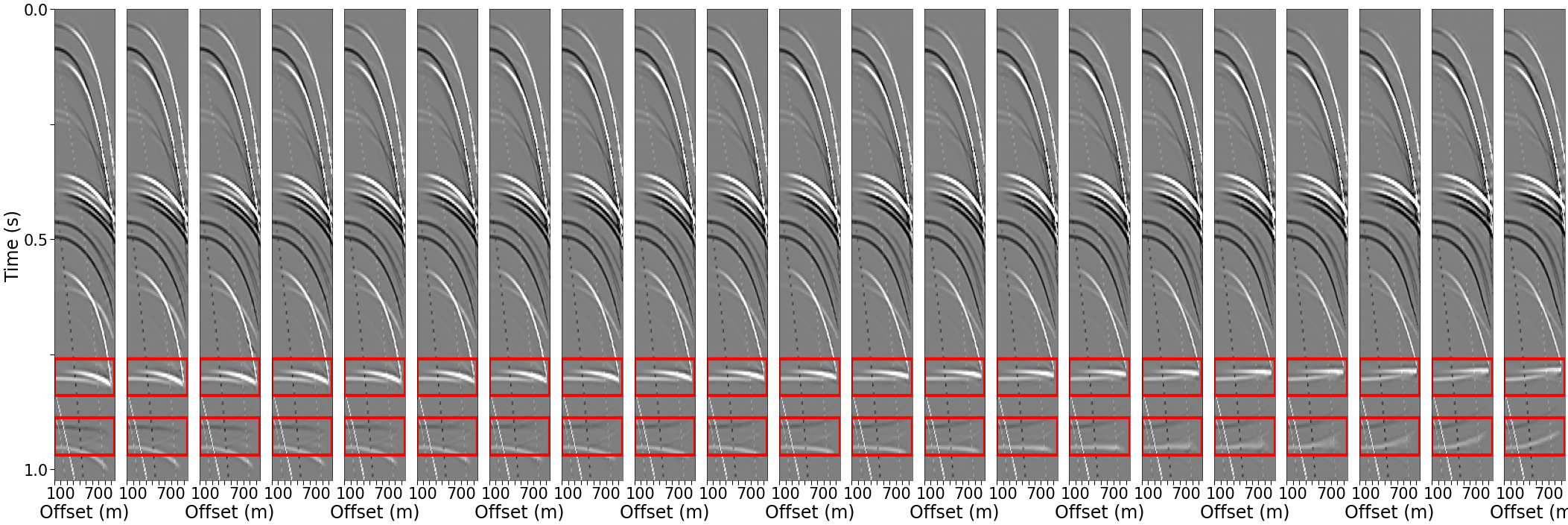}
    \caption{Predicted Multiples \model{}}
  \end{subfigure}
  \caption{Synthetic data results comparing our \unet{} baseline, and our \model{} model to the ground truth. Our model correctly removes the flat event around 0.8 seconds. The flat event should be removed because all the primaries are over-corrected, and the multiples therefore appear flat instead of downwards sloping.}
  \label{fig:synthetic_results_0}
\end{figure}

Figure \ref{fig:synthetic_results_0} illustrates a case where \model{} correctly identifies and removes a flat event at approximately 0.8 seconds that appears due to over-correction of primaries. The \unet{} baseline fails to recognize this and retains the event in the prediction.

\begin{figure}[htbp]
  \centering
  \begin{subfigure}[b]{\textwidth}
    \centering
    \includegraphics[width=\textwidth]{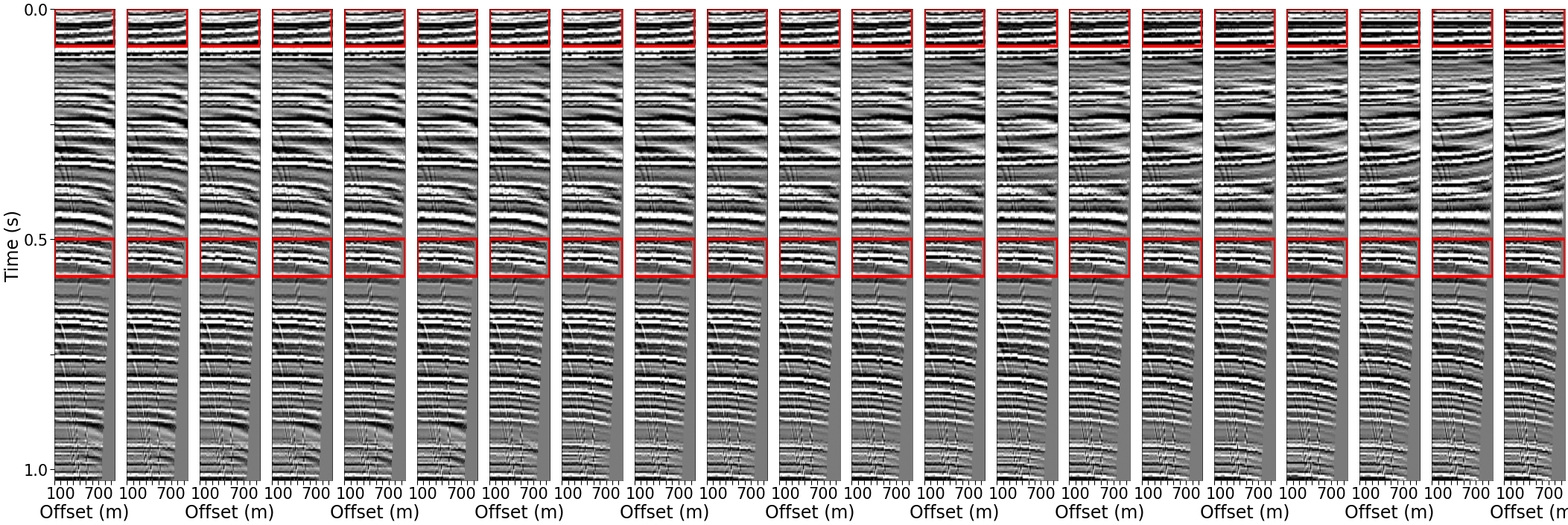}
    \caption{Gather}
  \end{subfigure}
  \hfill
  \begin{subfigure}[b]{\textwidth}
    \centering
    \includegraphics[width=\textwidth]{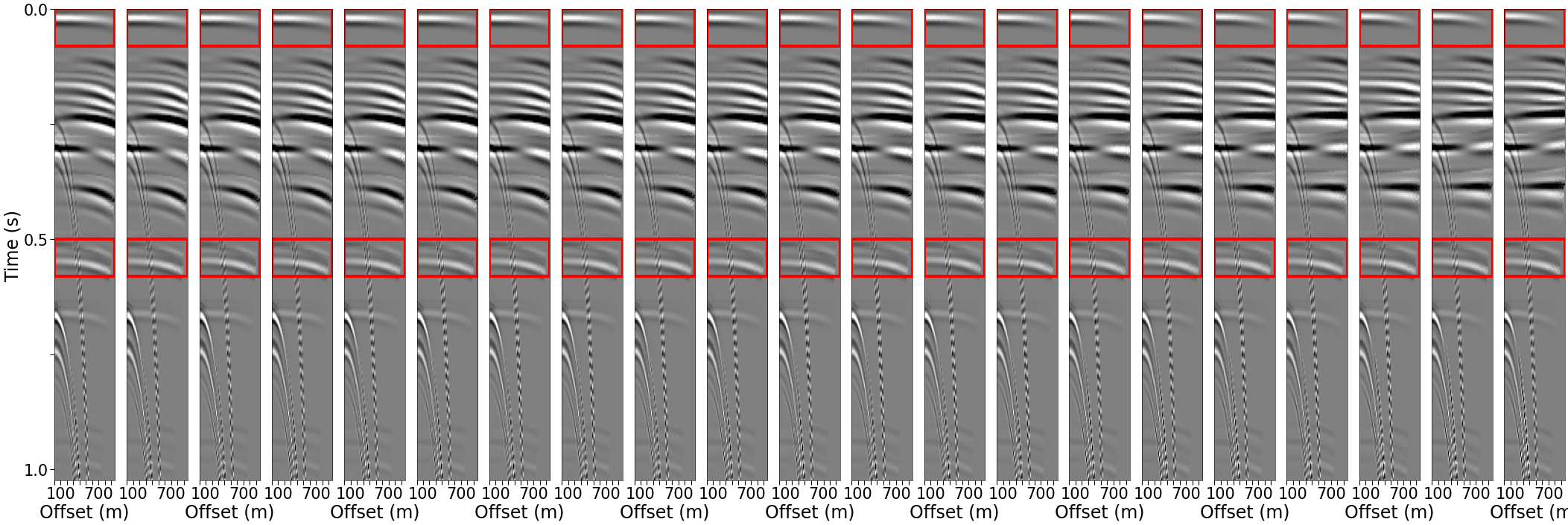}
    \caption{Multiples}
  \end{subfigure}
  \hfill
  \begin{subfigure}[b]{\textwidth}
    \centering
    \includegraphics[width=\textwidth]{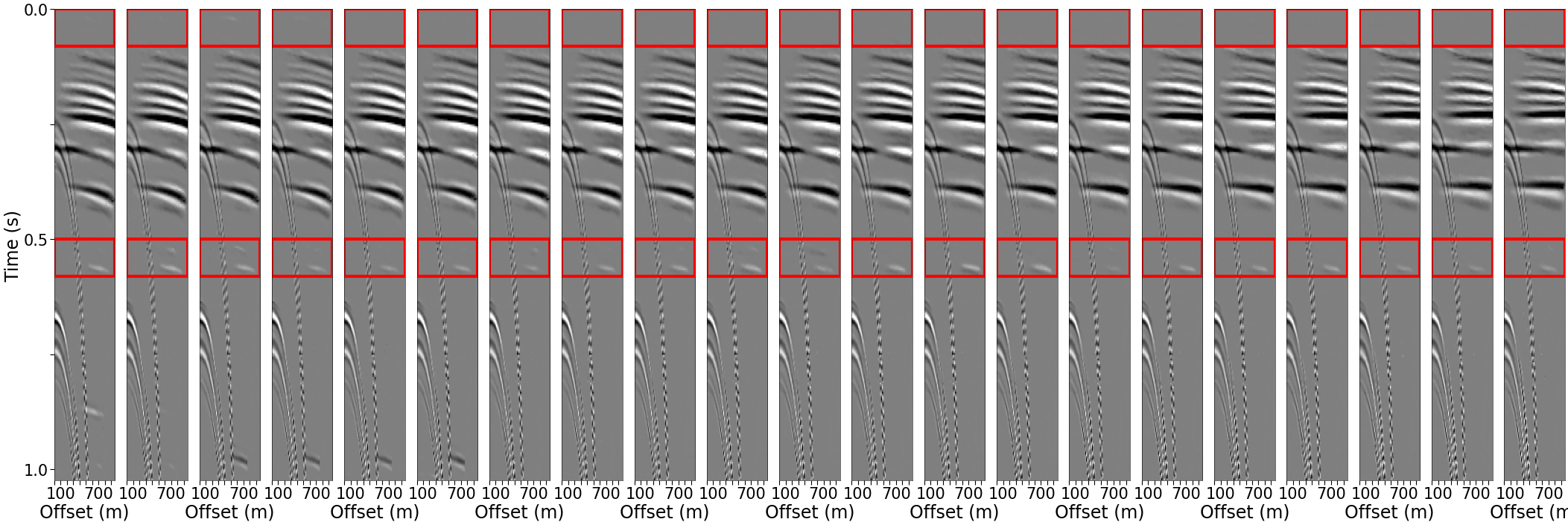}
    \caption{Predicted Multiples \unet{}}
  \end{subfigure}
  \hfill
  \begin{subfigure}[b]{\textwidth}
    \centering
    \includegraphics[width=\textwidth]{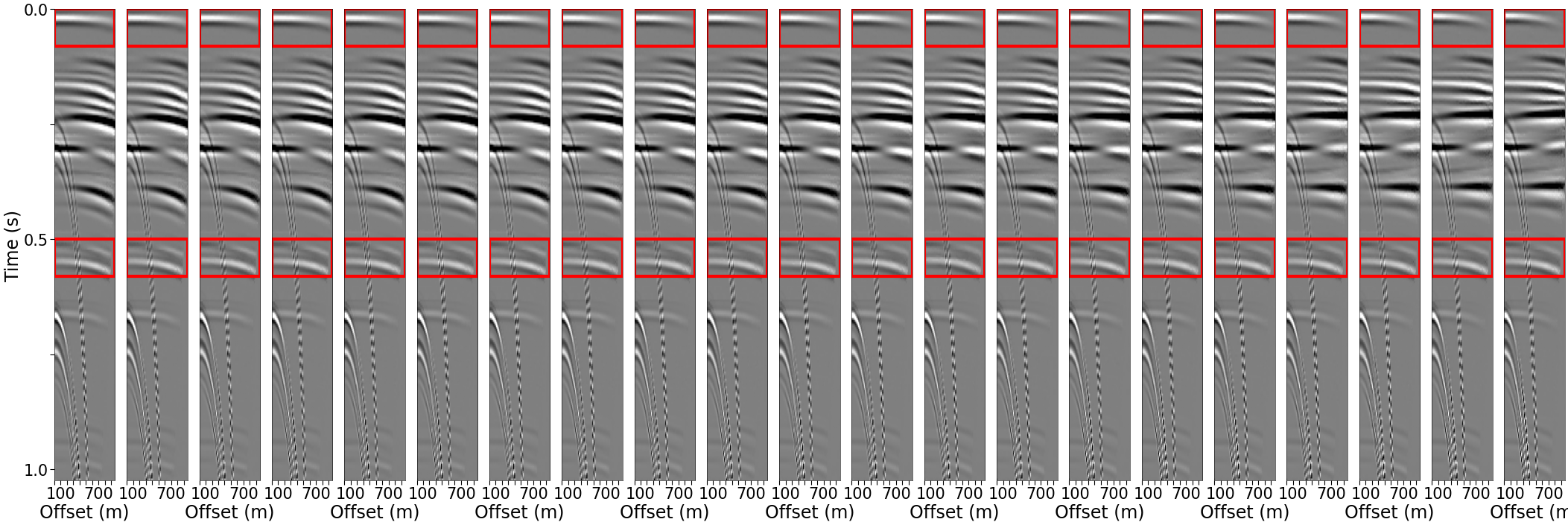}
    \caption{Predicted Multiples \model{}}
  \end{subfigure}
  \caption{Synthetic data results comparing our \unet{} baseline, and our \model{} model to the ground truth. Our model completely removes the event around 0.55 seconds across all \glspl{cdp}. Meanwhile the \unet{} baseline only partially removes the event and also not consistently across the whole line.}
  \label{fig:synthetic_results_13450}
\end{figure}

Figure \ref{fig:synthetic_results_13450} demonstrates \model{}'s ability to completely remove a multiple event at approximately 0.55 seconds across all \glspl{cdp}. In contrast, the \unet{} baseline achieves only partial removal with inconsistent performance across the seismic line.

\begin{figure}[htbp]
  \centering
  \begin{subfigure}[b]{\textwidth}
    \centering
    \includegraphics[width=\textwidth]{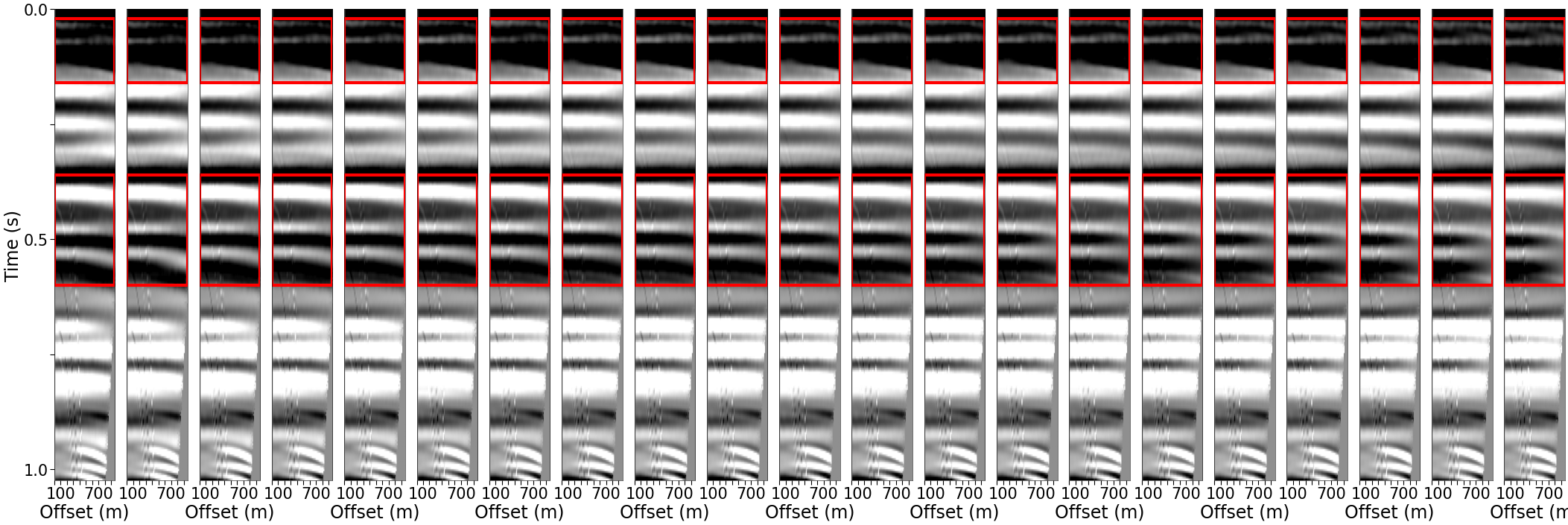}
    \caption{Gather}
  \end{subfigure}
  \hfill
  \begin{subfigure}[b]{\textwidth}
    \centering
    \includegraphics[width=\textwidth]{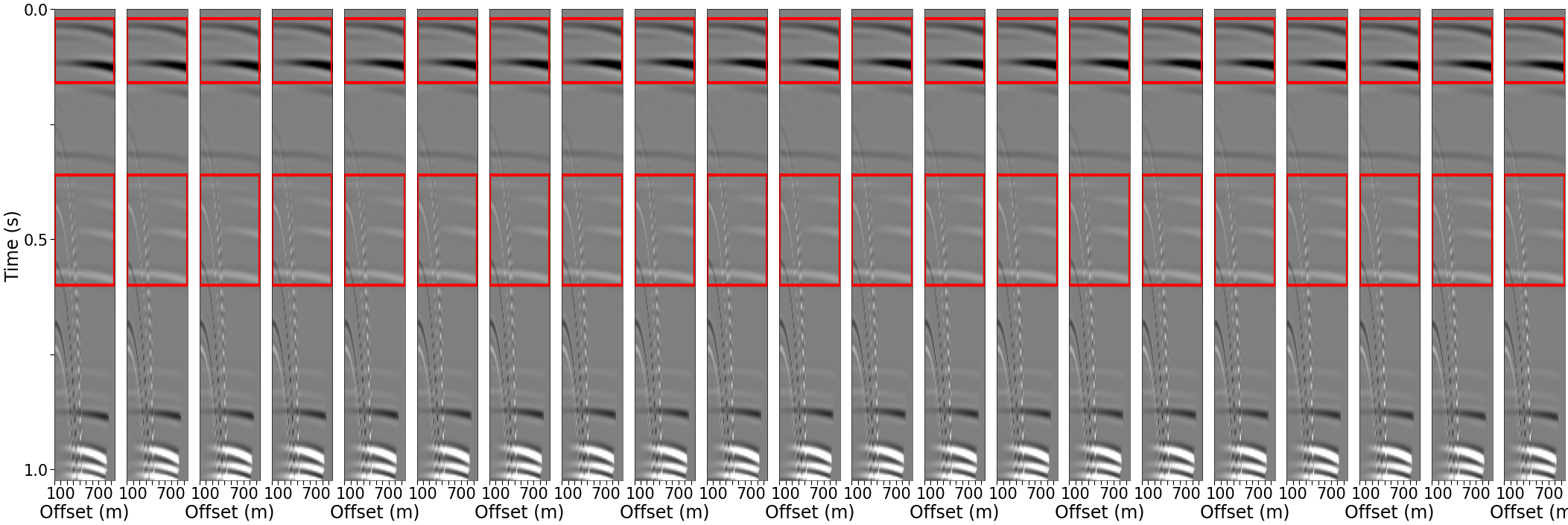}
    \caption{Multiples}
  \end{subfigure}
    \hfill
  \begin{subfigure}[b]{\textwidth}
    \centering
    \includegraphics[width=\textwidth]{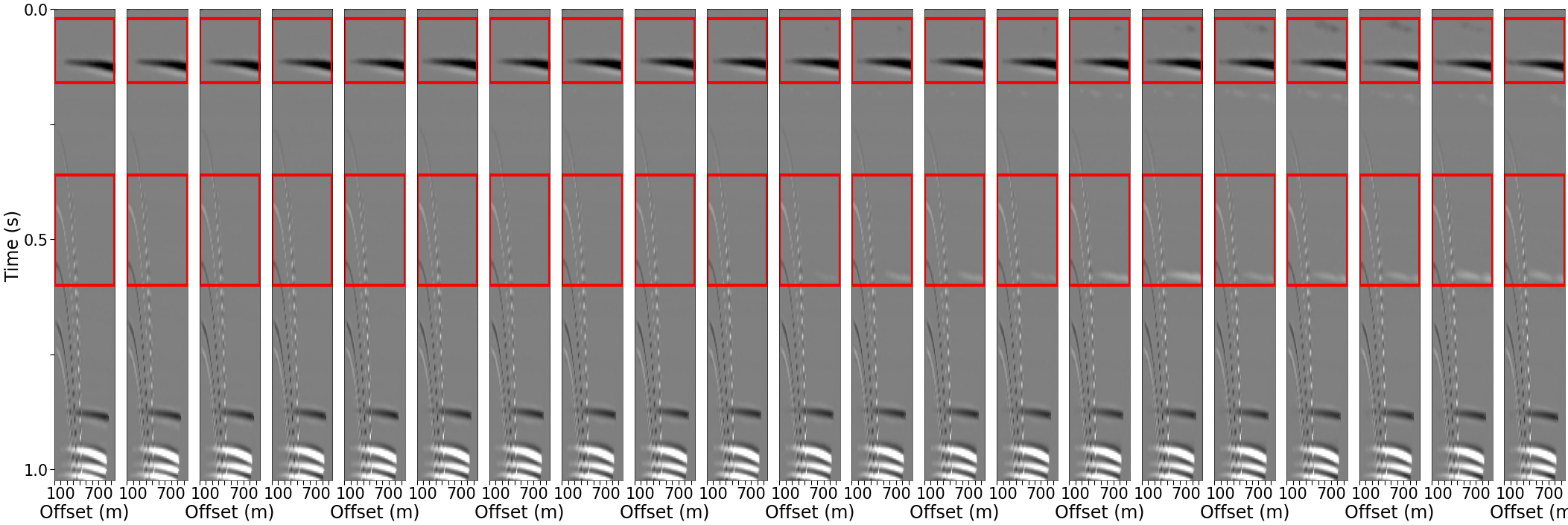}
    \caption{Predicted Multiples \unet{}}
  \end{subfigure}
  \hfill
  \begin{subfigure}[b]{\textwidth}
    \centering
    \includegraphics[width=\textwidth]{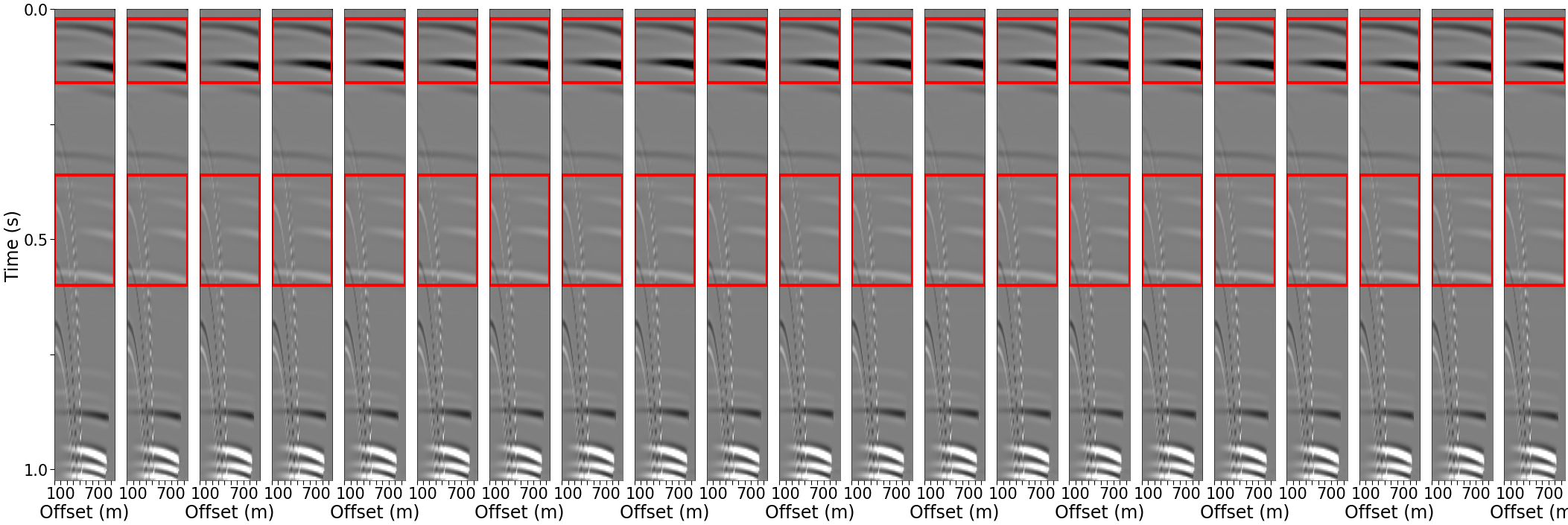}
    \caption{Predicted Multiples \model{}}
  \end{subfigure}
  \caption{Synthetic data results comparing our \unet{} baseline, and our \model{} model to the ground truth. Our model completely removes the event around 0.6 seconds across all \glspl{cdp}. Meanwhile the \unet{} baseline only partially removes the event and also only for the last few \glspl{cdp}.}
  \label{fig:synthetic_results_14950}
\end{figure}

Figure \ref{fig:synthetic_results_14950} shows similar behavior for an event at approximately 0.6 seconds. \model{} removes this event completely across all \glspl{cdp}, while the \unet{} baseline provides partial removal limited to the final \glspl{cdp} in the line.

\section{Improved Generalization}
\label{app:field_generalization}

Figure \ref{fig:field_generalization} examines generalization performance on field data. Subfigure \ref{fig:field_generalization_mul_onnx} shows results from a \unet{} trained on 100,000 gathers, demonstrating strong generalization. Subfigure \ref{fig:field_generalization_mul_no_prompting} presents results from a \unet{} trained on only 10,500 gathers, revealing limited generalization capability. Subfigures \ref{fig:field_generalization_mul_prompting} and \ref{fig:field_generalization_mul_prompting_onnx} display \model{} predictions using Radon-based prompts and prompts from the better trained \unet{} (shown in Subfigure \ref{fig:field_generalization_mul_onnx}), respectively. Despite training on only 10,500 gathers (same dataset than the \unet{} in Subfigure \ref{fig:field_generalization_mul_no_prompting}), \model{} maintains consistent generalization performance with both prompt types, matching the quality achieved by the \unet{} trained on ten times more data.

\begin{figure}[htbp]
  \centering
  \begin{subfigure}[b]{\textwidth}
    \centering
    \includegraphics[width=\textwidth]{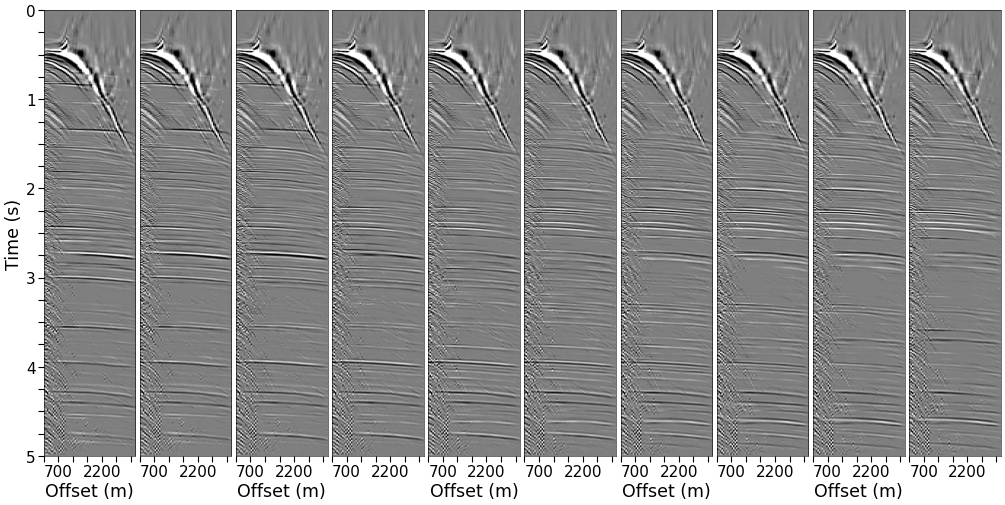}
    \caption{Multiples of a \unet{} trained with 100.000 gathers.}
    \label{fig:field_generalization_mul_onnx}
  \end{subfigure}
  \hfill
  \begin{subfigure}[b]{\textwidth}
    \centering
    \includegraphics[width=\textwidth]{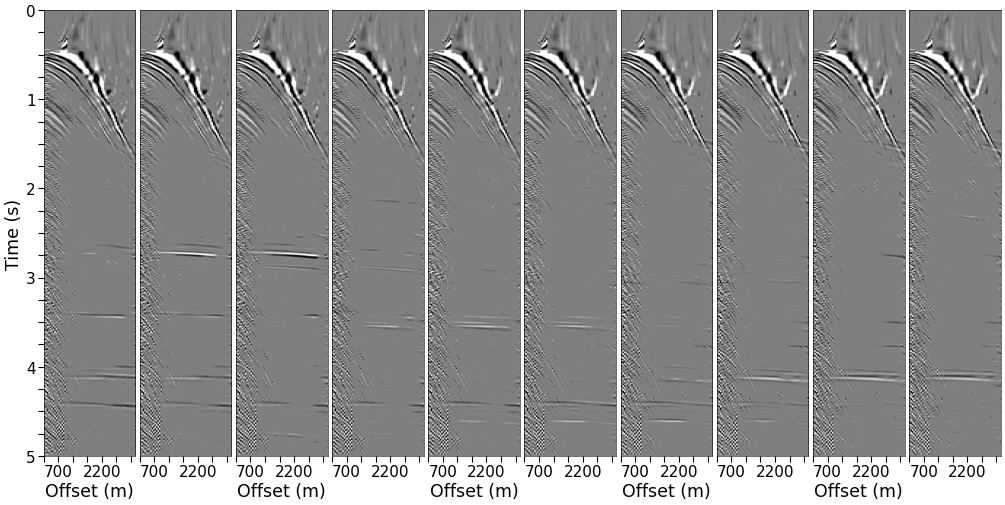}
    \caption{Multiples of \unet{} trained with 10.500 gathers.}
    \label{fig:field_generalization_mul_no_prompting}
  \end{subfigure}
\end{figure}

\begin{figure}[htbp]
  \ContinuedFloat
  \centering
  \begin{subfigure}[b]{\textwidth}
    \centering
    \includegraphics[width=\textwidth]{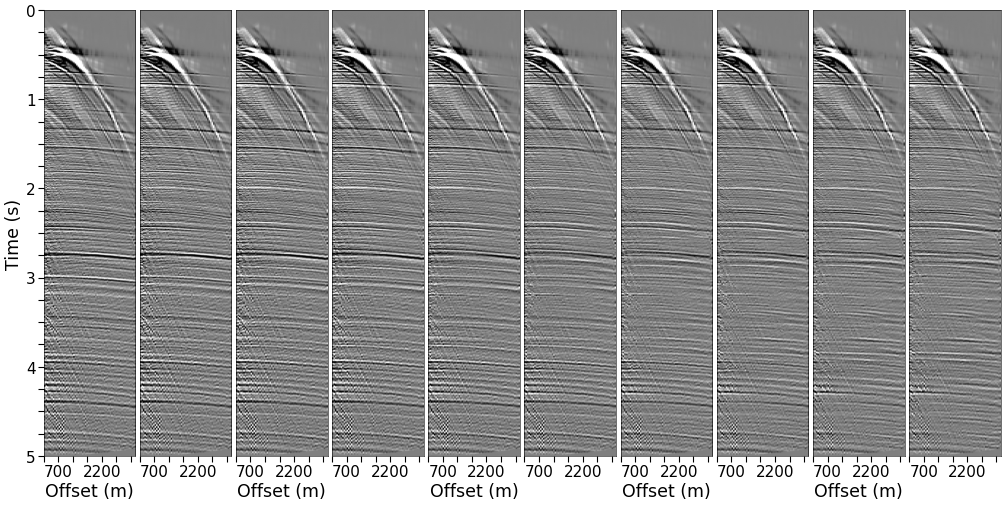}
    \caption{Multiples of \model{} trained with 10.500 gathers and prompted with Radon results}
    \label{fig:field_generalization_mul_prompting}
  \end{subfigure}
  \hfill
  \begin{subfigure}[b]{\textwidth}
    \centering
    \includegraphics[width=\textwidth]{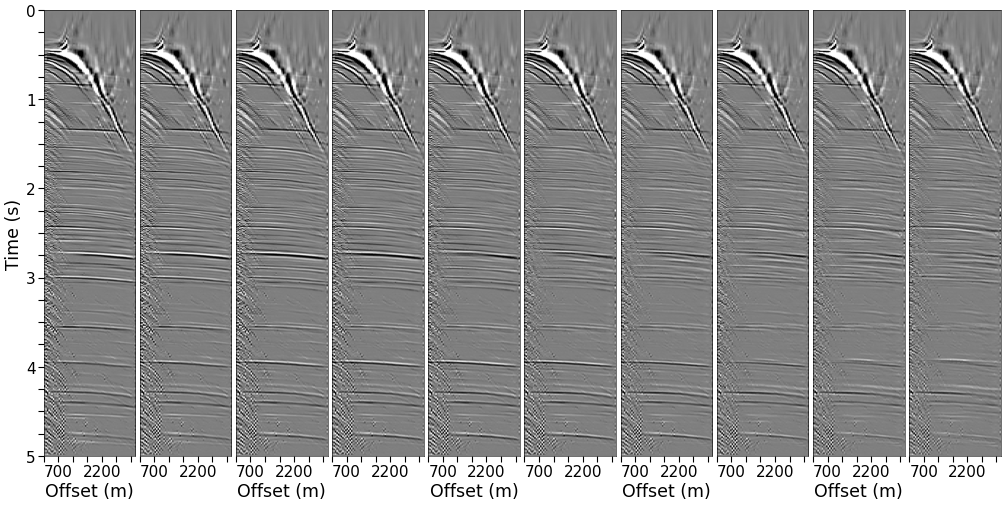}
    \caption{Multiples of \model{} trained with 10.500 gathers and prompted with results of the U-Net trained with 100.000 gathers.}
    \label{fig:field_generalization_mul_prompting_onnx}
  \end{subfigure}
  \caption{Performance comparison of two \unet s trained on datasets of different sizes (100,000 vs. 10,500 gathers) and \model{} with Radon and \unet{} prompts. The \unet{} trained on 100,000 gathers demonstrates effective generalization, while the model trained on 10,500 gathers (used for synthetic data) shows limited generalization capability. In comparison, \model{} maintains consistent generalization performance with both prompt types.}
  \label{fig:field_generalization}
\end{figure}

\end{document}